\documentclass[11pt,a4paper]{article}

\usepackage[T1]{fontenc}
\usepackage[utf8]{inputenc}
\usepackage{lmodern}
\usepackage{microtype}
\usepackage[a4paper,margin=2.1cm]{geometry}
\usepackage{amsmath,amssymb,amsfonts}
\usepackage{graphicx}
\usepackage{xcolor,colortbl}
\usepackage{booktabs,tabularx,array,multirow}
\usepackage{float}
\usepackage{placeins}
\usepackage{adjustbox}
\usepackage[ruled,vlined]{algorithm2e}
\usepackage{siunitx}
\usepackage[font=small,labelfont=bf]{caption}
\usepackage[font=small,labelfont=bf]{subfig}
\usepackage[numbers,sort&compress]{natbib}
\usepackage{authblk}
\usepackage[hidelinks]{hyperref}

\newcolumntype{C}{>{\centering\arraybackslash}X}
\title{\bfseries Protect the Brain When Treating the Heart: Feasibility of 2.5D U-Net for Real-Time Gaseous Microemboli Detection}
\author[1,3,*]{Andrea Angino}
\author[2]{Ken Trotti}
\author[5,6,7,*]{Diego Ulisse Pizzagalli}
\author[1,2]{Rolf Krause}
\author[4]{Tiziano Torre}
\author[4,5,8]{Stefanos Demertzis}

\affil[1]{Universit\`a della Svizzera Italiana (USI), Center for Computational Medicine in Cardiology (CCMC), Lugano, Switzerland; \href{mailto:andrea.angino@usi.ch}{andrea.angino@usi.ch} (A.A.); \href{mailto:rolf.krause@usi.ch}{rolf.krause@usi.ch} (R.K.)}
\affil[2]{King Abdullah University of Science and Technology (KAUST), AMCS/CEMSE Division, Thuwal, Saudi Arabia; \href{mailto:ken.trotti@kaust.edu.sa}{ken.trotti@kaust.edu.sa} (K.T.)}
\affil[3]{UniDistance Suisse, Brig, Switzerland}
\affil[4]{Cardiac Surgery Department, Cardiocentro Ticino Institute, Ente Ospedaliero Cantonale (EOC), Lugano, Switzerland; \href{mailto:tiziano.torre@eoc.ch}{tiziano.torre@eoc.ch} (T.T.); \href{mailto:stefanos.demertzis@eoc.ch}{stefanos.demertzis@eoc.ch} (S.D.)}
\affil[5]{Faculty of Biomedical Sciences, Universit\`a della Svizzera Italiana (USI), Lugano, Switzerland; \href{mailto:diego.pizzagalli@usi.ch}{diego.pizzagalli@usi.ch} (D.U.P.)}
\affil[6]{Theodor Kocher Institute, Faculty of Medicine, University of Bern, Bern, Switzerland}
\affil[7]{MeDiTech institute, Department of Innovative Technologies, SUPSI, Lugano, Switzerland}
\affil[8]{Faculty of Medicine, University of Bern, Bern, Switzerland}
\affil[*]{Correspondence: \href{mailto:andrea.angino@usi.ch}{andrea.angino@usi.ch} (A.A.); \href{mailto:diego.pizzagalli@usi.ch}{diego.pizzagalli@usi.ch} (D.U.P.)}
\date{}

\hypersetup{
  pdftitle={Protect the Brain When Treating the Heart: Feasibility of 2.5D U-Net for Real-Time Gaseous Microemboli Detection},
  pdfauthor={Andrea Angino, Ken Trotti, Diego Ulisse Pizzagalli, Rolf Krause, Tiziano Torre, Stefanos Demertzis},
  pdfkeywords={computer vision; biomedicine; echocardiography; gaseous microemboli; U-Net; real-time segmentation; cardiac surgery}
}

\begin{document}
\maketitle

\begin{abstract}
Gaseous microemboli (GME) represent a common complication of cardiac structural interventions across both surgical and transcatheter approaches.
Intraoperative transesophageal echocardiography (TEE) represents a convenient methodology to monitor and visualize the presence of circulating GME.
However, their detection and quantification are far from trivial due to operator-dependent view, high velocity, and objects with similar structure in the background.
Here, we propose a feasibility study based on a 2.5D U-Net architecture to detect GME in space-time connected data.
We applied and tested such an architecture on a pilot dataset of eight TEE recordings ($60$ fps, $600\times 800$ pixels) from eight different patients undergoing cardiac surgery, resulting in improved detection of moving GMEs against the background with respect to classical spot detection algorithms and 2D U-Net, yet retaining real-time execution speed with respect to more complex deep-learning architectures.
Under leave-one-patient-out cross-validation, the selected model achieved strong detection performance under a three-pixel radius-tolerant grace-zone evaluation, with a precision of 92.55\% and recall of 80.54\%, corresponding to radius-tolerant Intersection over Union (IoU) and Dice coefficients of 73.95\% and 84.13\%, respectively. Complementarily, strict pixel-based segmentation metrics were also computed, yielding an IoU of 41.74\% and a Dice coefficient of 57.98\%.
The selected model achieved an average inference time of $0.12 s$ per batch on the tested hardware. To assess specificity on unseen data, we additionally evaluated the model on an external GME-negative TEE dataset, where it produced predominantly empty or near-empty masks, indicating a low rate of spurious detections.
These results support the technical feasibility of real-time GME segmentation, although broader clinical validation on larger multi-patient, multicenter datasets is still required.
\end{abstract}


\section{Introduction}

Cardiac structural interventions (surgical and/or transcatheter) are associated with significant neurological risks due to the inadvertent introduction of emboli - solid or gaseous - into the bloodstream. Gaseous microemboli (GME), composed of environmental air, are especially concerning as they can travel through the circulatory system with a slow reabsorption rate, potentially lodging in cerebral vessels and leading to silent infarctions and cognitive deficits~\cite{ChungEM}. Neurological complications arising from GME have been documented extensively, ranging from transient impairments to long-term cognitive decline in patients following cardiac procedures~\cite{Kihara2021}. Existing solutions, such as transcranial Doppler ultrasound and diffusion-weighted MRI, provide a means of postoperative emboli detection, yet they lack the immediacy necessary to facilitate real-time intervention during surgery, limiting their effectiveness in mitigating intraoperative neurological risks~\cite{Orihashi2024}. Providing real-time feedback within the surgical workflow can therefore support the operating team by offering timely confidence estimates and improving intraoperative decision-making~\cite{Kurz2022}.

Cardiac echocardiography stands as an effective procedure that can be easily integrated into clinical practice to detect GMEs in cardiac chambers. Among the imaging modalities, trans-esophageal echocardiography (TEE) is particularly convenient for intraoperative monitoring. However, reliable software for automatic GME detection and quantification is still lacking. This gap is largely explained by the small size of emboli, their irregular and variable appearance, and their strong similarity to surrounding tissue and common echocardiographic artifacts, which makes purely appearance-based detection unreliable.

The contribution of this work is threefold: first, we describe a dedicated annotation workflow for GME detection in intraoperative echocardiographic videos. Second, we adapt a lightweight 2.5D U-Net to exploit short-term temporal information while preserving real-time feasibility. Third, we integrate the segmentation output into a prototype monitoring pipeline that provides frame-wise GME area estimates. Hence, we provide a technical feasibility study on a pilot dataset, that could set the basis for a more complete clinical validation and a support to the doctors during image segmentation to build the dataset.

\section{Related Work}\label{S:related_work}

Learning-based segmentation provides a natural way to address these challenges. State-of-the-art architectures such as Fully Convolutional Networks (FCNs)~\cite{Long_2015_CVPR} and Mask R-CNN~\cite{He_2017_ICCV} have been successfully applied to medical image segmentation, while U-Net remains a widely adopted reference model due to its ability to preserve fine spatial details through skip connections~\cite{ronneberger2015unet}. Beyond static segmentation, recent work has shown that U-Net-based models can also segment motile targets in grayscale images with limited resolution~\cite{Antonello2023}.
However, when the input consists of video or temporally structured imaging data, relying on single-frame appearance alone can be limiting, especially when the target objects are small, rapidly moving, and visually similar to background artifacts. In such cases, short-term temporal information can provide discriminative cues that are not available from individual frames. This motivates the use of 2.5D approaches, in which a short stack of adjacent frames or slices is provided as input, offering a practical compromise between purely 2D models and computationally heavier 3D architectures.

In the specific case of GME, appearance alone is often insufficient as emboli may be indistinguishable from static speckle patterns and transient artifacts in single frames, while their rapid motion across consecutive frames provides a key cue for discrimination. Building on this evidence, we propose a real-time detection system based on a lightweight 2.5D U-Net architecture for GME segmentation in echocardiographic video. The model processes short sequences of consecutive frames to introduce limited temporal context while remaining computationally feasible for online deployment under real-time intraoperative constraints.

\section{Dataset construction}\label{S:Dataset}

Segmenting GME requires meticulous, time-intensive work by trained specialists capable of distinguishing microemboli from surrounding cardiac structures.

Segmentation in this context presents unique challenges: while it may initially seem straightforward to delineate hyperechogenic regions within each frame, the task is complicated by the fact that cardiac muscle and emboli often share similar grayscale intensities on echocardiographic evaluation. This similarity creates ambiguity, especially near edges, where motion is the key indicator distinguishing rapidly moving emboli from the slower-moving or stationary cardiac tissue. Manual analysis typically requires moving through video frames to observe these subtle differences in movement, as GMEs are characterized by rapid positional changes compared to the relatively stable cardiac background.
\begin{figure}
    \centering
    \includegraphics[width=0.8\linewidth]{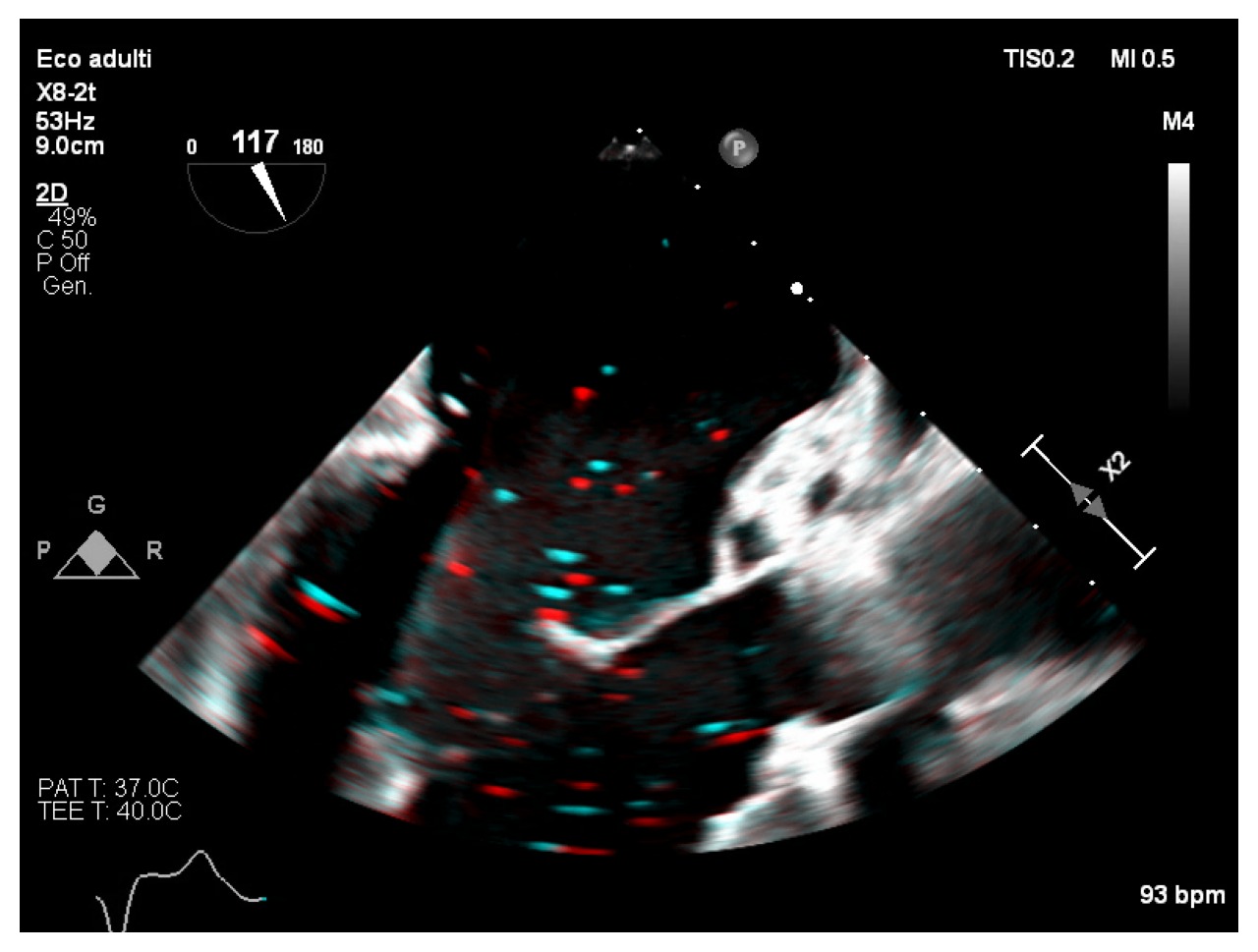}
    \caption{
    Color-coded overlay of two consecutive frames: the initial frame is depicted in red, while the subsequent frame appears in turquoise. Regions that remain stationary across the two frames appear white, corresponding to the stable cardiac background. The distinct red/turquoise displacement highlights the dynamic movement of GMEs.}
    \label{fig:imm1}
\end{figure}
Figure~\ref{fig:imm1} presents two overlapping subsequent frames, with the first shown in red and the second in turquoise. This color-coded overlay highlights motion cues, enabling the precise identification of emboli against the stationary cardiac background.

For this study, we developed a custom annotation tool tailored to meet the specific requirements of GME segmentation in echocardiographic frames. This tool provides intuitive navigation through video frames, allowing expert annotators to segment emboli using two selection modes: a manual outline tool for precise boundary tracing and an automatic region selection tool that identifies regions based on color variation. Although the automatic mode expedites the segmentation process, it can occasionally lack precision, making manual corrections necessary. The tool also includes user-friendly features, such as an “undo” function to reverse recent actions, a “cancel” option for discarding wrongly selected regions, and an auto-save feature to ensure that all changes are securely stored. The annotation tool and supporting software resources are openly available in the AirCatch GitHub repository~\cite{Aircatch}. The repository includes installation instructions, dependency information, and a Quick Start example to facilitate the use of the software.

Annotations were generated with a custom tool for manual annotation, and clinically reviewed under the supervision of the cardiac surgery specialists among the authors (T.T. and S.D.). Ambiguous regions were assessed by inspecting consecutive frames, since motion relative to the cardiac background was the main criterion for identifying GME. A post-hoc independent annotation reproducibility analysis was subsequently performed on a randomly selected subset of the dataset and is reported in Section~\ref{sec:postInterObs}.

Although frame subdivision increases the number of localized training samples, the dataset remains limited in terms of independent clinical acquisitions.

\begin{table}[H]
\caption{Dataset summary and split information used for final model performance evaluation.\label{tab:dataset_summary}}
\begin{tabularx}{\textwidth}{lC}
\toprule
\textbf{Variable} & \textbf{Value}\\
\midrule
Number of echocardiographic videos & 8\\
Number of patients/procedures & 8\\
Approximate duration per video & 2 s\\
Frame rate & 60 frames/s\\
Physical area per pixel (mean $\pm$ SD mm$^2$/pixel) & $0.0886 \pm 0.031 $ \\
Frame resolution & 600 $\times$ 800 pixels\\
Patch size after subdivision & 300 $\times$ 400 pixels\\
Approximate number of patches & $\sim$4000\\
Split strategy & Leave-one-patient-out cross-validation\\
\bottomrule
\end{tabularx}
\end{table}

Because of the limited number of source videos, the present results should be interpreted as pilot within-dataset feasibility results rather than as proof of patient-independent generalization, which will require larger and more diverse multi-patient datasets.

\subsection{Post-hoc Inter-Observer Agreement Analysis}
\label{sec:postInterObs}

To provide a quantitative assessment of annotation reproducibility, $30$ frames were randomly selected from the annotated dataset and independently re-annotated by a clinician in training who had not previously seen either the dataset or the original annotations. The re-annotation was performed with access to adjacent frames but without access to the reference masks. The original consensus annotations are denoted by (A), while the independent re-annotations are denoted by (B).

Individual GMEs were extracted from each binary mask using connected-component analysis. The centroid of each connected component was defined as the mean (x)- and (y)-coordinates of all pixels belonging to that object. Objects in (A) and (B) were matched one-to-one according to the Euclidean distance between their centroids, using a maximum matching distance of $5$ pixels.

\begin{figure}[!htbp]
    \centering
    \subfloat[Centroid-based agreement.]{%
        \includegraphics[width=0.48\linewidth]{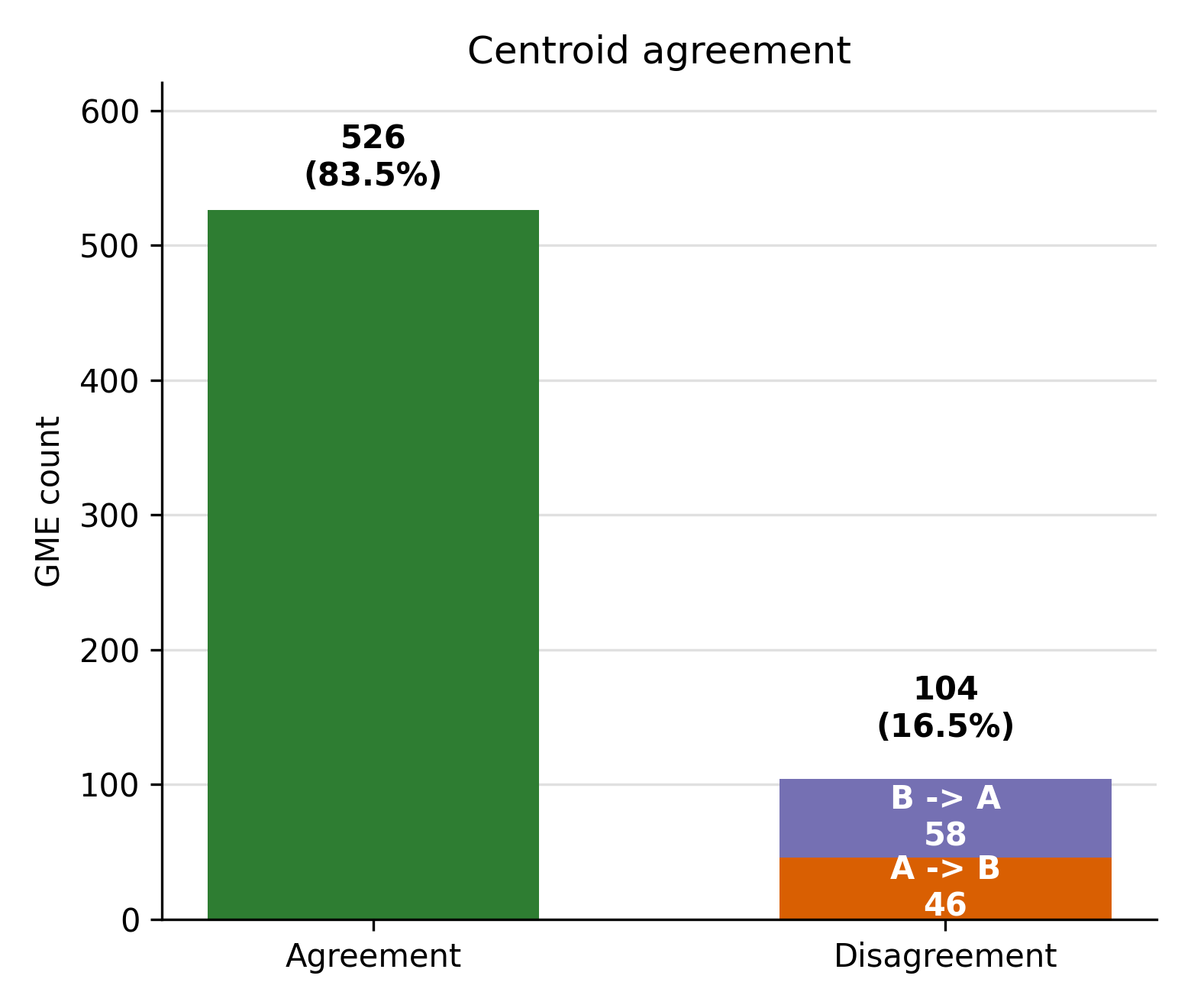}%
    }
    \hfill
    \subfloat[GME count correlation.]{%
        \includegraphics[width=0.48\linewidth]{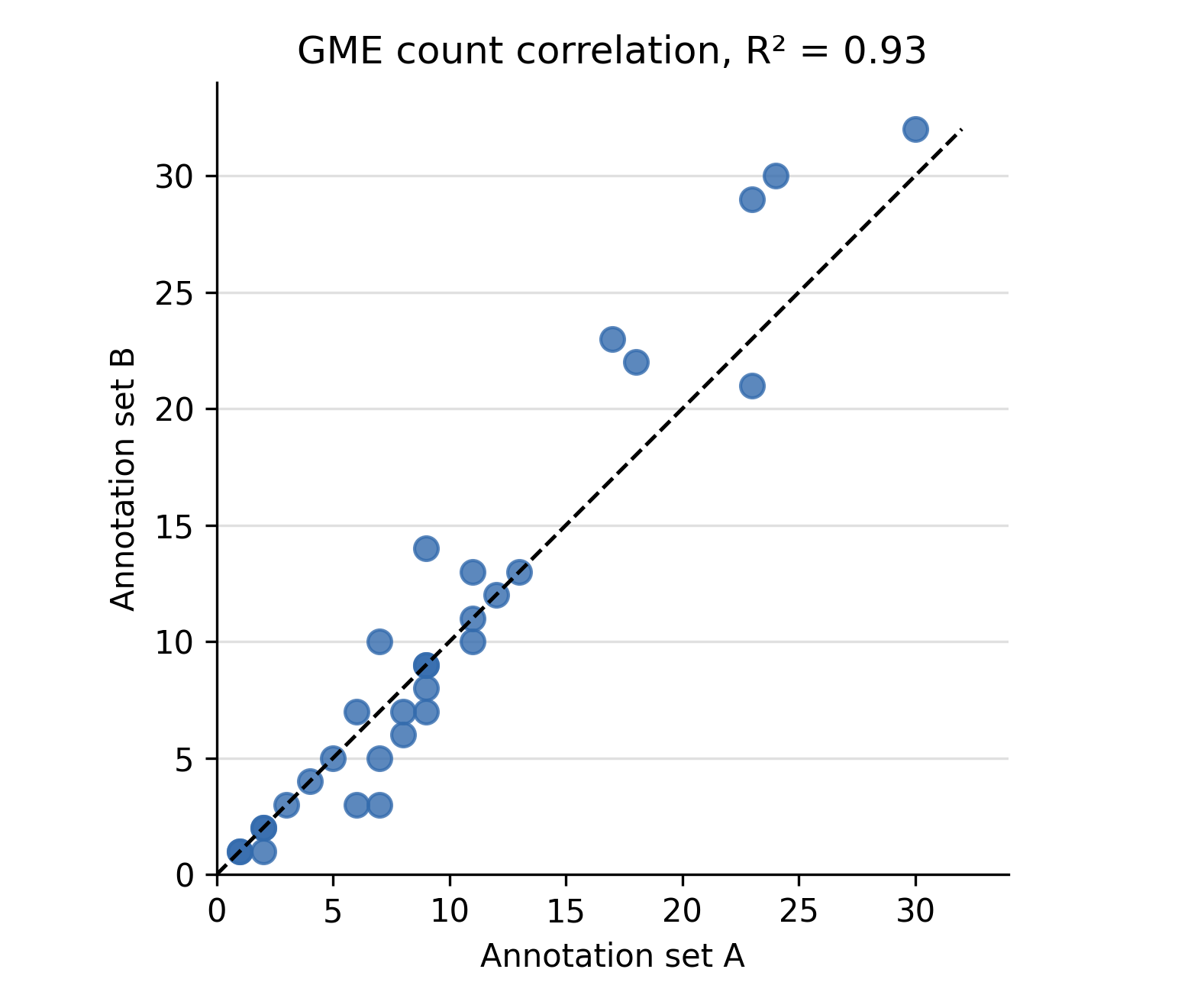}%
    }
    \caption{Post-hoc inter-observer analysis on $30$ randomly selected frames. Panel (a) shows object-level agreement between the original consensus annotations (A) and the independent re-annotations (B), based on one-to-one centroid matching with a maximum $5$ pixels distance. Panel (b) shows the frame-wise correspondence between the numbers of GMEs identified in (A) and (B). Observed correlation $R^2=0.93$.}
    \label{fig:InterObs}
\end{figure}

Across the $30$ re-annotated frames, 526 GME objects were matched between the two annotation sets, while 46 objects in (A) and 58 objects in (B) had no corresponding match. In addition, the frame-wise numbers of annotated GMEs showed strong correspondence between the two annotation sets, with $R^2=0.93$, as shown in Fig.~\ref{fig:InterObs}

This analysis indicates reproducibility in identifying and spatially localizing individual GME objects. It should nevertheless be distinguished from pixel-wise contour reproducibility. Qualitative inspection of matched annotations showed that independent observers could identify the same GME while assigning different boundaries. Such differences are particularly relevant for small objects represented at limited spatial resolution, since TEE provides a two-dimensional representation of a moving three-dimensional GME. Consequently, small variations in boundary placement may substantially affect strict pixel-wise overlap even when object identity and localization are concordant. These observations provide additional empirical support for considering the proposed radius-tolerant metrics introduced in Section~\ref{sec:metrics} as complementary measures.

\section{Neural network structure}\label{S:NNstructure}
\begin{figure}[!htbp]
\centering
\subfloat[Intensity-based thresholding.]{\includegraphics[width=0.45\linewidth]{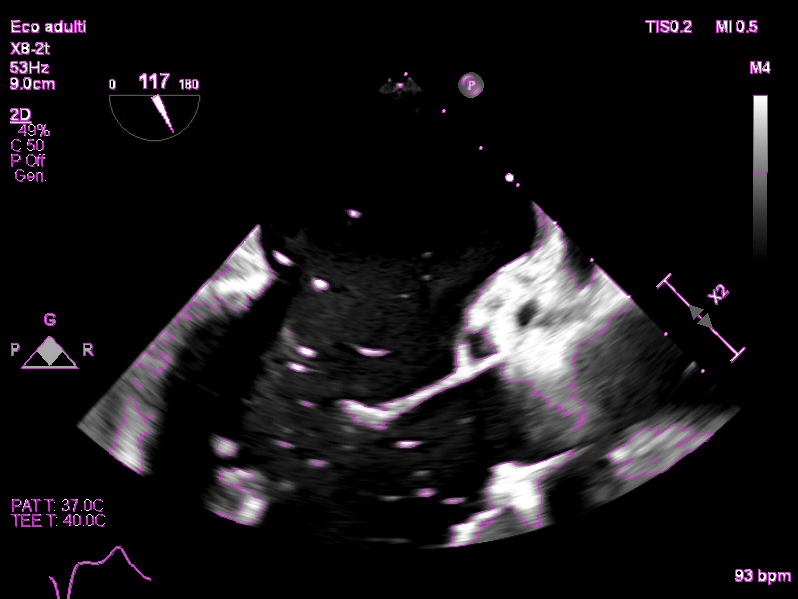}}
\hfill
\subfloat[LoG-based spot detection.]{\includegraphics[width=0.45\linewidth]{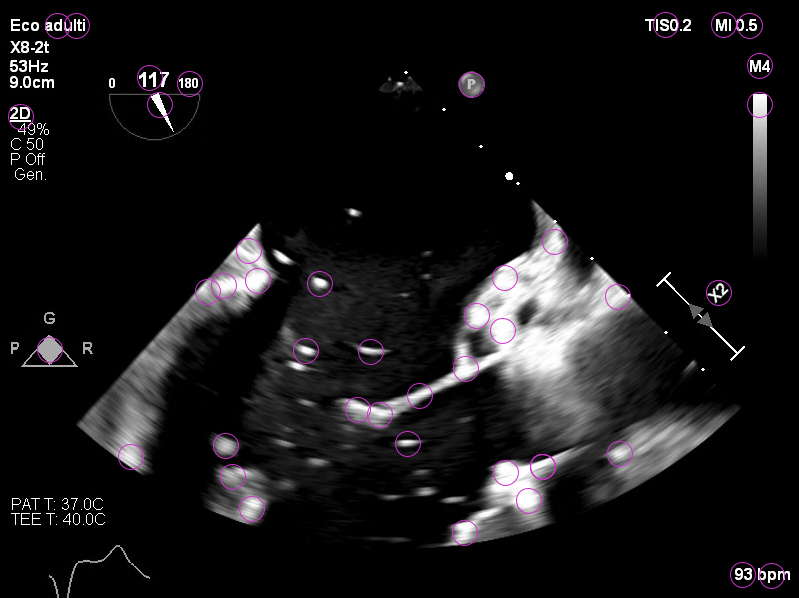}}

\medskip
\subfloat[2D U-Net segmentation.]{\includegraphics[width=0.45\linewidth]{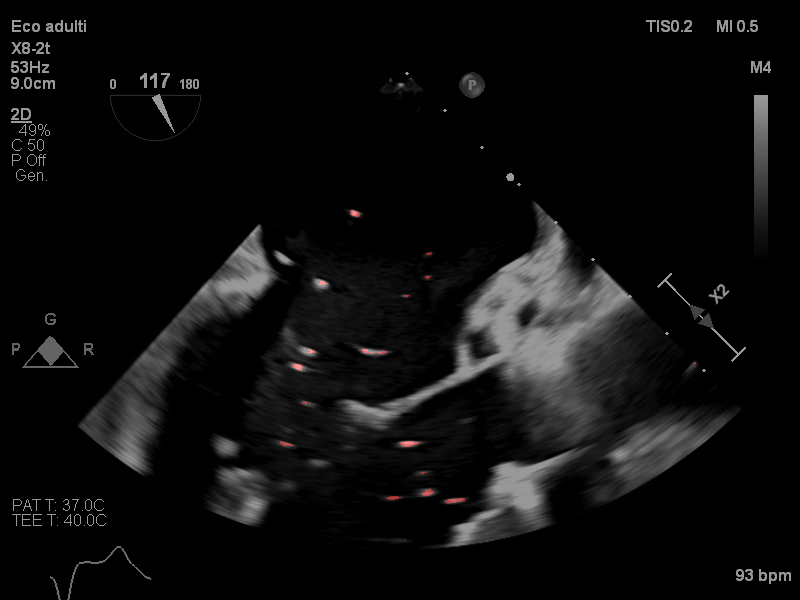}}
\hfill
\subfloat[2.5D U-Net segmentation.]{\includegraphics[width=0.45\linewidth]{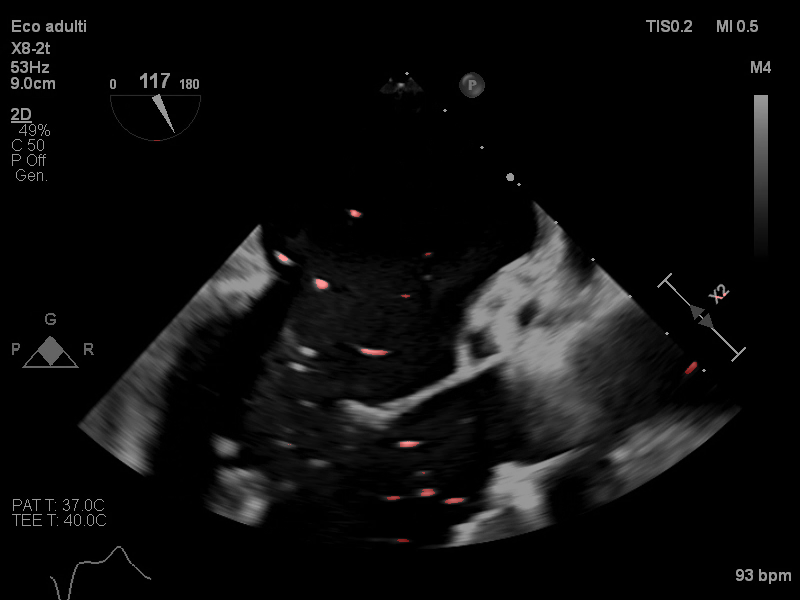}}

\medskip
\subfloat[Manually segmented ground truth.]{\includegraphics[width=0.45\linewidth]{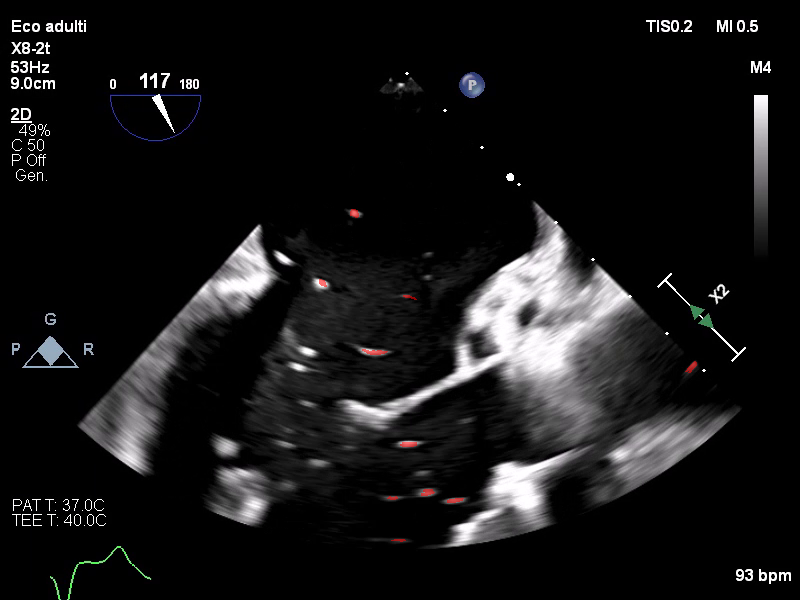}}
\caption{Representative qualitative comparison of segmentation results obtained using classical image-processing methods (a,b), deep learning-based models (c,d), and manually segmented ground truth (e). In panels (a--d), colored overlays indicate the regions segmented as GMEs by each method, superimposed on the original echocardiographic frame. Panel (e) shows the manually annotated ground-truth GME regions using the same overlay convention.}
\label{fig:segmentation_comparison}
\end{figure}

To provide a comparative baseline, we first evaluated classical segmentation approaches using the open-source Fiji/TrackMate 7~\cite{Ershov2022} software (Thresholding and Laplacian of Gaussian (LoG) filtering strategies). The top-left and top-right panels in Figure~\ref{fig:segmentation_comparison} correspond to the segmentation outputs obtained with these methods.
As can be seen, both approaches capture not only the GMEs but also portions of the surrounding muscular tissue, resulting in over-segmentation. The LoG method shows improved sensitivity to GMEs but still fails to suppress responses from cardiac structures with similar intensity and texture characteristics.

These limitations of traditional image processing approaches highlighted the need for a learning-based strategy capable of capturing more complex spatial and contextual features, motivating our decision to adopt a deep learning framework.

A 2D U-Net architecture, which is widely recognized for its effectiveness in biomedical image segmentation~\cite{ronneberger2015unet}, struggles to effectively distinguish between GME and the surrounding cardiac boundaries due to the limited size of our dataset and the intrinsic complexity of differentiating microemboli from surrounding tissue structures.

Considering the importance of motion cues in differentiating emboli from background tissue, an observation made during dataset construction, the main improvement in model architecture was achieved by transitioning to a 2.5D U-Net~\cite{Unet25D}. Unlike a 3D U-Net, which processes a full volumetric space-time block of data, and hence is not suitable for real-time processing, a 2.5D U-Net partially incorporates the temporal dimension by including a short sequence of frames along the time axis.
This allows the network to capture essential temporal information while remaining computationally efficient for real-time applications. As shown in Figures~\ref{fig:segmentation_comparison}.c and~\ref{fig:segmentation_comparison}.d, comparing with the manually segmented reference in ~\ref{fig:segmentation_comparison}.e, the 2D model often misclassifies background resembling GMEs and overestimates the true GMEs area, whereas the 2.5D model excludes the background.

The choice of a U-Net architecture stems from its well-documented success in biomedical imaging tasks due to its encoder-decoder structure~\cite{Unet2}, which enables detailed segmentation by leveraging contextual information. The U-Net’s skip connections allow the model to retain spatial information lost during downsampling, crucial for accurately segmenting GME boundaries against a complex background.

All patients undergoing cardiac surgery are equipped with a direct invasive arterial line, a central venous catheter, and a bladder catheter. Under conditions of general anesthesia and orotracheal intubation, a trans-esophageal ultrasound probe is positioned, which allows continuous monitoring of cardiac contractility, the functionality of the heart valves, and the result of the operation performed.
The video frames were acquired by the echocardiographic machine (Philips Epiq) during the weaning phases from the CardioPulmonary Bypass (CPB) when any bubbles present inside the cardiac cavities are pushed forward by the resumption of pulmonary ventilation.
To enable real-time processing and sharing, an HDTV video capture USB 3.0 device was used to connect the Philips Epiq to a laptop, mirroring the echocardiographic images while capturing high-definition video for NN analysis. The NN processes batches of
10 frames, introducing a delay of 0.16 s for image loading and batch formation during which the previous batch is processed, and a batch inference time of 0.12 s, ensuring near real-time visualization with a bounded overall delay of approximately 0.28 s. This setup allows doctors to perform their assessments as if the AI were not present, relying solely on the live echocardiographic feed. However, they can also refer to the laptop screen to visualize quantitative bubble estimates and the regions identified by the neural network as potential GMEs, providing an additional layer of insight.

The binary segmentation procedure, described in Algorithm~\ref{alg:unet}, takes a raw grayscale frame as input and applies a pretrained U-Net to produce a binary mask. The input frame is first divided into four quadrants, which are processed in batches by the network. The resulting per-pixel probabilities are reassembled into a full-frame probability map and thresholded to obtain the final binary segmentation.
\begin{algorithm}[!htbp]
\small
\caption{2.5D U-Net segmentation with frame length history $lh$}
\label{alg:unet}
\KwIn{Frames $F \in \mathbb{R}^{lh\times 600 \times 800}$, pretrained 2.5D U-Net $\mathcal{U}$, threshold $t \in [0,1]$, EMA parameter $\alpha \in [0,1]$}.
\KwOut{Binary mask $M\in {\{0,1\}}^{600 \times 800}$, smoothed GME area estimate $\bar{A}$}
\SetKwBlock{Pre}{Preprocess}{}
\Pre{
  Partition $F$ into patches $\{F_k\}_{k=1}^4$\tcp*{$F_k\in\mathbb{R}^{lh\times300\times400}$}\label{line:split}
  Form the tensor $\mathbf{X}=[F_1,F_2,F_3,F_4]$ \tcp*{$X\in\mathbb{R}^{4\times lh\times300\times400}$}\label{line:stack}
}
\SetKwBlock{Infer}{Network Inference}{}
\Infer{
  $\mathbf{P} \leftarrow \mathcal{U}(\mathbf{X})$\tcp*{$\mathbf{P}\in[0,1]^{4\times 300\times 400}$}
  Aggregate probability patches into $P\in[0,1]^{600\times 800}$\;\label{line:reassemble}
}
\SetKwBlock{Thresh}{Thresholding}{}
\Thresh{
  \ForEach{pixel $P_{i j}$ in $P$}{
    \uIf{$P_{i j} \ge t$}{
      $M_{i j} \leftarrow 1$\tcp*{foreground}
    }
    \Else{
      $M_{i j} \leftarrow 0$\tcp*{background}
    }
  }
}
\SetKwBlock{Post}{Post-processing}{}
\Post{
  $A \leftarrow \sum_{i,j} M_{ij}$\tcp*{Raw predicted GME area in pixels}
  $\bar{A} \leftarrow \alpha A + (1-\alpha)\bar{A}_{prev}$\tcp*{EMA-smoothed displayed area}
}
\Return $M$, $\bar{A}$\tcp*{Binary mask and smoothed area estimate}
\end{algorithm}
\paragraph{Hardware requirements}
For practical real-time deployment of the proposed parallel processing pipeline, we recommend a system equipped with an $8$ GB or larger CUDA-enabled GPU and an $8$-core processor. This configuration should be regarded as a practical deployment recommendation rather than an experimentally established minimum hardware threshold.

This setup ensures energy and resource efficiency and can be physically located in the operating room near the echocardiography machine. The machine requires multiple cores to distribute the workload across several processing pipelines.

One pipeline is dedicated to the NN's inference, another is responsible for computing the estimate by processing the segmented data through an exponential moving average, effectively prioritizing recent changes while retaining long-term trends for accuracy, and a third for generating the real-time output, which results from combining the previous contributions into a coherent visualization (See example in Figure~\ref{fig:realtime_segmentation}).
This division of tasks ensures seamless integration and uninterrupted real-time operation, even in resource-constrained environments.

\subsection{Segmentation Performance Metrics}\label{sec:metrics}
To comprehensively evaluate segmentation performance, we consider three complementary metrics: the Frequency Bias Index (FBI), the Intersection over Union (IoU), and the Dice coefficient. Each captures different aspects of segmentation quality, allowing for a nuanced assessment of model behavior in detecting GME regions.\\
Throughout this section, we use the following notation:
\begin{itemize}
  \item \(\mathrm{TP}\) (true positives): number of GME pixels correctly predicted as GME,
  \item \(\mathrm{FP}\) (false positives): number of background pixels incorrectly predicted as GME,
  \item \(\mathrm{FN}\) (false negatives): number of GME pixels missed by the model,
  \item \(\mathrm{TN}\) (true negatives): number of background pixels correctly predicted as background.
\end{itemize}
\paragraph{Frequency Bias Index}
The Frequency Bias Index (FBI) is used in this work as a primary calibration metric for estimating the overall GME burden. Since the clinical objective is to quantify the total embolic area visible in the echocardiographic stream, FBI provides a direct measure of whether the predicted GME extent is systematically under- or over-estimated. It is commonly used in forecasting to measure the tendency of a model to over- or under-predict positive events. In our context, it is defined as the ratio between the number of pixels predicted as GME and the total number of ground-truth GME pixels, rescaled to a percentage for visualization purposes:

\begin{equation}\label{eq:FBI}
\text{FBI} = \frac{\text{TP} + \text{FP}}{\text{TP} + \text{FN}} \times 100 =\left( 1 + \frac{\text{FP} - \text{FN}}{\text{TP} + \text{FN}} \right) \times 100.
\end{equation}

A value close to 100$\%$ indicates that the predicted segmentation has approximately the same area as the ground-truth GME, regardless of its spatial alignment. For this reason, FBI should be interpreted as a quantitative area-estimation metric rather than as a measure of spatial localization accuracy. Spatial agreement between prediction and annotation is therefore evaluated separately through IoU and Dice metrics. The right-hand side of Equation~\ref{eq:FBI} highlights how the FBI reflects the net discrepancy between predicted and true GME regions: if $\text{FN} > \text{FP}$, the model underestimates the GME extent and the FBI falls below 100$\%$. Conversely, if $\text{FP} > \text{FN}$, the model overestimates it and the FBI exceeds 100$\%$.

This property makes the FBI particularly suitable for our application, where capturing the full extent of the GME is critical, as a slight overestimation (Type I error dominant, $\text{FP}>\text{FN}$), which may lead to unnecessary deairing procedures, is more acceptable than underestimation (Type II error dominant, $\text{FN}>\text{FP}$), which could result in failing to intervene on patients at risk due to high GME levels. Thus, a model with an FBI slightly above 100$\%$ is preferable to one below 100$\%$, as it reflects a conservative yet balanced estimate of the GME area.

\paragraph{Intersection over Union (IoU)}
The Intersection over Union (IoU), or Jaccard index, quantifies the overlap between predicted and true GME regions in terms of pixel counts. It is defined as
\begin{equation}\label{eq:IoU_TPFPFN}
\mathrm{IoU}
\;=\;
\frac{\mathrm{TP}}{\mathrm{TP} + \mathrm{FP} + \mathrm{FN}}
\times 100\%
\end{equation}
An IoU of 100\% means perfect agreement, while 0\% means no overlap. Because every FP and FN reduces the score equally, IoU is a balanced measure of segmentation quality.

\paragraph{Dice Coefficient}
The Dice coefficient (also called the F$_1$-score for segmentation) emphasizes the harmonic mean of precision and recall. In terms of pixel counts:
\begin{equation}\label{eq:Dice_TPFPFN}
\mathrm{Dice}
\;=\;
\frac{2\,\mathrm{TP}}{2\,\mathrm{TP} + \mathrm{FP} + \mathrm{FN}}
\times 100\%
\end{equation}
Dice ranges from 0\% (no overlap) to 100\% (perfect overlap). By giving double weight to TP, Dice is more sensitive to small structures and less punishing of isolated FP or FN than IoU.

Consequently, model selection was based on the combined interpretation of FBI, IoU, and Dice, with particular emphasis on FBI due to the quantitative embolic-load estimation objective of the study.

\paragraph{Radius-Tolerant Grace-Zone Evaluation}\label{sec:grace_zone_metrics}
The main objective of this study is reliable GME detection and embolic-load estimation rather than exact contour agreement at single-pixel resolution. Therefore, in addition to strict pixel-wise segmentation metrics described above, we introduce a radius-tolerant grace-zone evaluation that tolerates small contour errors around annotated GME masks, focusing on GME detection. This is appropriate for our application because, in this formulation, predictions falling close to the annotated GMEs boundaries are treated as acceptable detections rather than hard segmentation errors.

\begin{figure}[!htbp]
    \centering
\includegraphics[width=0.8\linewidth]{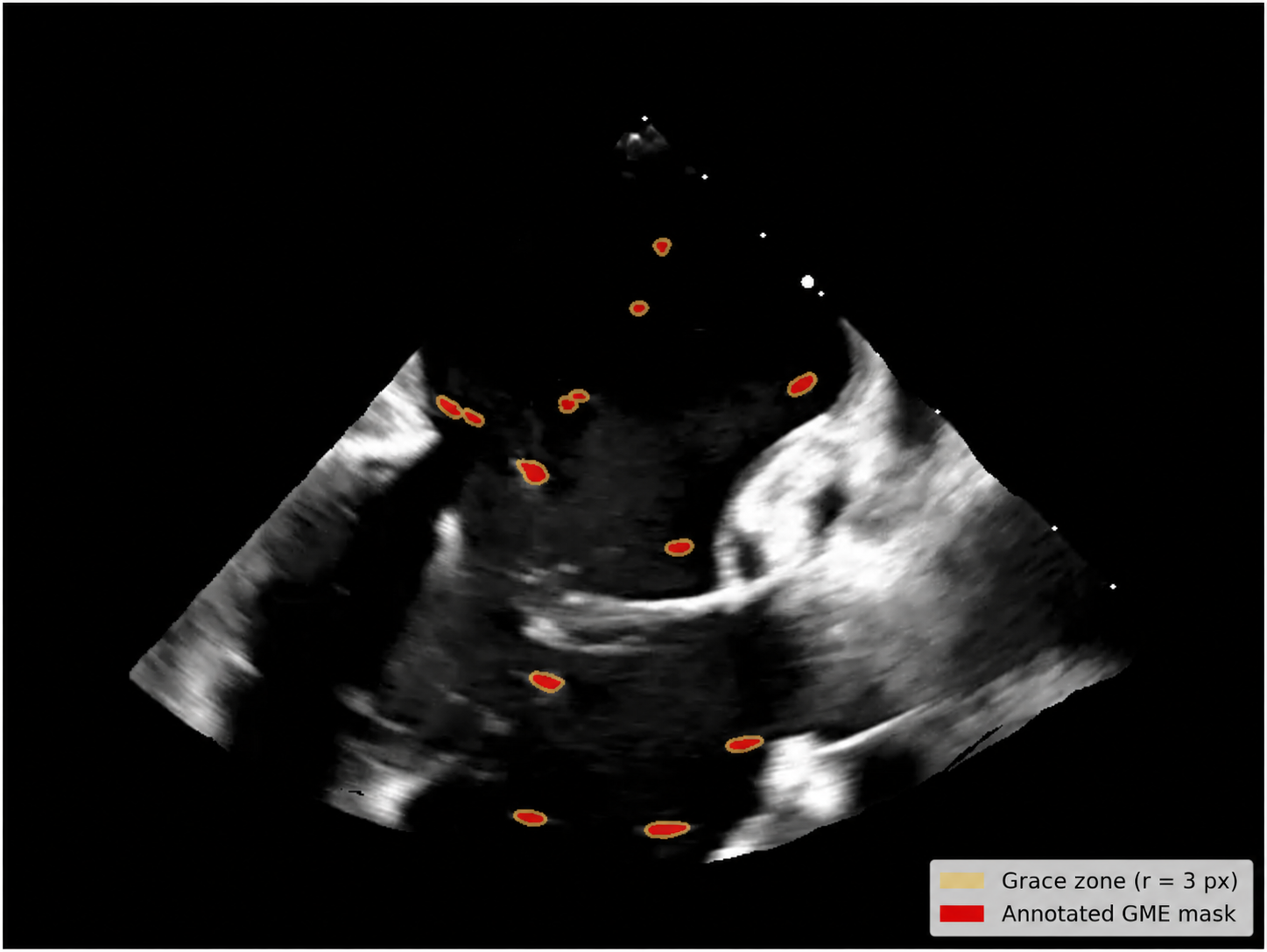}
    \caption{Illustration of the radius-tolerant grace-zone evaluation with a grace-zone radius of $r = 3$ px. Red regions represent the annotated GME mask, while the yellow bands correspond to the tolerance regions generated by the structuring element $D_r$. Predicted foreground pixels falling inside these regions are treated as partially correct detections and therefore contribute to the effective true-positive term instead of being counted as strict false positives. The same tolerance principle is applied symmetrically to the prediction mask through $\hat{Y}_r$.}
    \label{fig:grace_zone_example}
\end{figure}

The radius-tolerant metrics provide a complementary assessment of GME
detection in the presence of small boundary ambiguities.

Let $Y \subseteq \Omega$ denote the ground-truth mask, let $\hat{Y} \subseteq \Omega$ denote the predicted mask, and let $D_r$ be the discrete disk of radius $r$ pixels. To account for boundary uncertainty, we define radius-tolerant neighborhoods around both the annotation and the prediction as
\begin{equation*}
Y_r = Y \oplus D_r,
\qquad
\hat{Y}_r = \hat{Y} \oplus D_r.
\end{equation*}
Under the following formulation, the radius-tolerant confusion terms are defined as

\begin{equation*}
FP_r^{\mathrm{eff}}
=|\hat{Y} \setminus Y_r|,
\qquad
FN_r^{\mathrm{eff}}
=|Y \setminus \hat{Y}_r|,
\end{equation*}

\begin{equation*}
TP_r^{\mathrm{eff}}
=\frac{1}{2}\left(
TP_r^{\mathrm{pred}}+TP_r^{\mathrm{gt}}\right).
\quad
TN_r^{\mathrm{eff}}
=|\Omega \setminus (Y_r \cup \hat{Y})|,
\end{equation*}
where
\begin{equation*}
TP_r^{\mathrm{pred}}
=
|\hat{Y} \cap Y_r|,
\qquad
TP_r^{\mathrm{gt}}
=
|Y \cap \hat{Y}_r|.
\end{equation*}

IoU, Dice, precision, recall, and F$_1$ are then computed from these radius-tolerant counts and denoted with subscript $r$. In this way, small boundary mismatches are penalized less severely, while clear false detections in safe background and missed GME cores remain penalized.

An example of this strategy is illustrated in Fig.~\ref{fig:grace_zone_example} with $r=3$, where the yellow regions correspond to the tolerance bands generated by $D_r$ around the annotated GMEs. Predicted foreground pixels falling inside these regions are treated as partially correct detections and therefore contribute to the effective true-positive term rather than being counted as strict false positives. The same reasoning is applied symmetrically to the prediction mask through $\hat{Y}_r$.

Tolerance-based evaluation is common in boundary-aware and medical-image segmentation, where small localization errors are accepted when predicted and reference contours fall within a prescribed distance \citep{arbelaez2011contour,cheng2021boundary,nikolov2021clinically}.
Following this idea, we define a radius-tolerant mask evaluation in which predicted positives within a radius-$r$ neighborhood of the ground-truth GMEs mask are counted as valid detections, while unpredicted pixels inside this neighborhood are treated as ambiguous rather than rewarded as true negatives.

The sensitivity of the resulting metrics to the choice of the radius $r$ is evaluated empirically in Section~\ref{sec:numres}.

\section{Numerical Results}\label{S:numerical_results}
This section presents all experimental results. In Section~\ref{sec:history}, we investigate the effect of input length, specifically, the number of frames provided to the 2.5D U-Net. Section~\ref{sec:tradeoff} presents a parameter study used to finalize the model configuration. For both Sections~\ref{sec:history} and~\ref{sec:tradeoff}, the dataset was split into $80\%$ training and $20\%$ testing sets for model selection and hyperparameter analysis. Section~\ref{sec:numres} reports the performance of the selected model evaluated using leave-one-patient-out cross-validation.

All experiments in Sections~\ref{sec:history}–\ref{sec:numres} were conducted under the same training conditions, as detailed below.
All models were trained for 50 epochs with a batch size of 10, using binary cross-entropy loss and Adam optimizer. The learning rate was set to $\mathrm{lr}=10^{-3}$ for the first 24 epochs and then reduced to $\mathrm{lr}=10^{-4}$ thereafter, using the default momentum values of $\beta_{1}=0.9$ and $\beta_{2}=0.999$. Results are averaged over three runs with different random initializations. Experiments were conducted on a Raider~A18~HX~A9W laptop (AMD Ryzen~9~9955HX3D CPU, 64\,GB RAM, NVIDIA RTX~5080 GPU with 16\,GB VRAM). Section~\ref{S5:negative_dataset} evaluates the models on the GME-negative SR-TEE dataset~\cite{Xu2023SRTEE} to assess spurious detections.

Finally, Section~\ref{S5.4:realtime} presents a qualitative real-time validation of the final system integrated into the online inference pipeline.

\subsection{Impact of Model History Length}\label{sec:history}
In the first experiment, we evaluate the impact of input sequence length on the performance of a 2.5D U-Net, specifically whether more than two consecutive frames offer any benefit. Due to delay constraints associated with longer histories, we limit the maximum history to three frames.
\begin{figure}[htbp]
  \centering
  \includegraphics[width=0.8\linewidth]{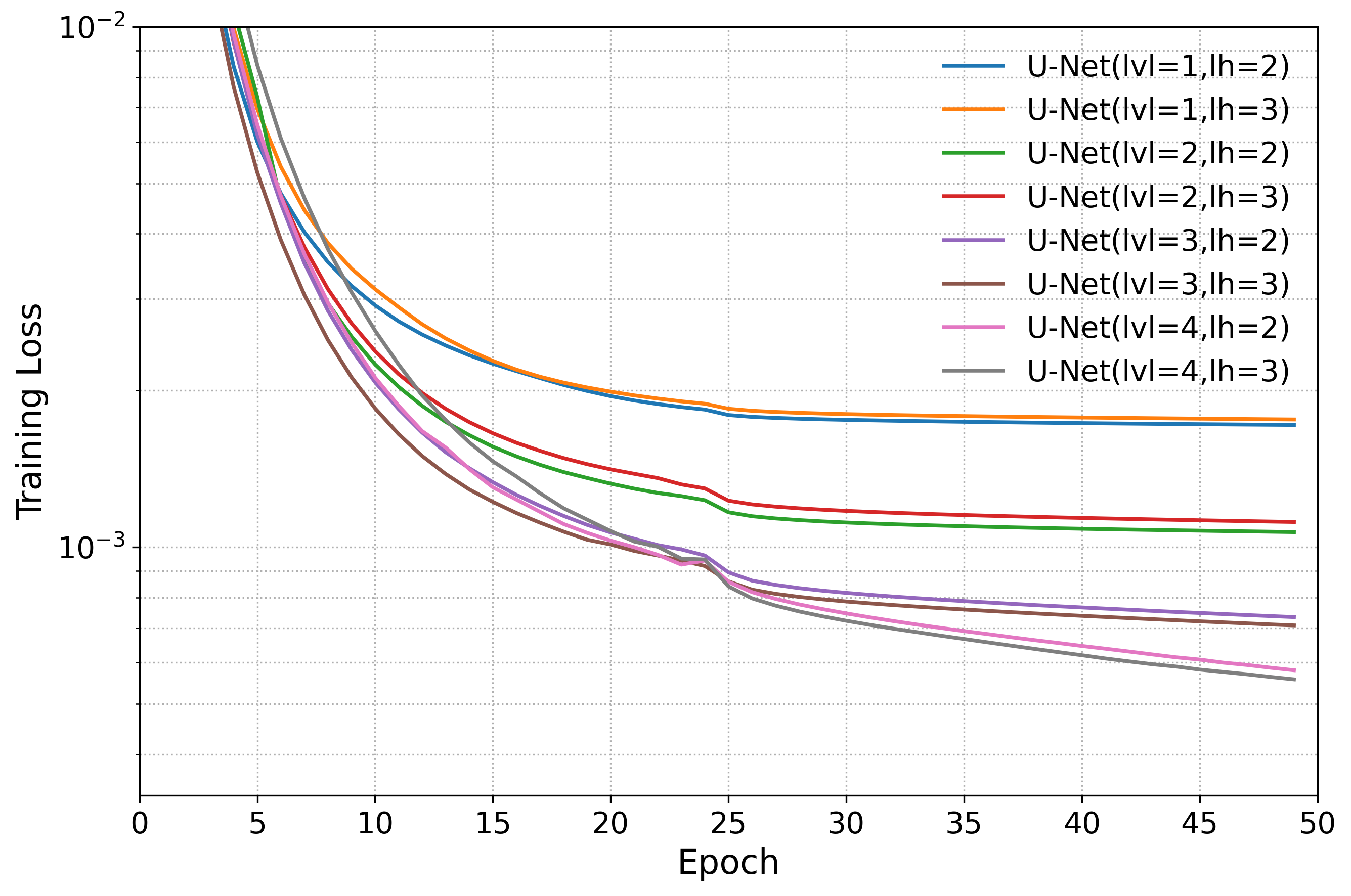}
  \caption{Training loss over 50 epochs for 2.5D U-Net models with 8 base channels, varying the number of down-sampling levels ($lvl \in \{1, 2, 3, 4\}$) and the input history length ($lh \in \{2, 3\}$). All models were trained using binary cross-entropy loss and the Adam optimizer with a batch size of 10. }
  \label{fig:train_loss}
\end{figure}

Figure~\ref{fig:train_loss} presents the training loss curves as a function of epochs for a 2.5D U-Net for different parameters. Specifically, the base number of feature map channels is set to 8 as we vary the number of downsampling levels ($\mathrm{lvl}\in\{1,2,3,4\}$) and the input history length ($\mathtt{lh}\in\{2,3\}$).

As expected, increasing the number of downsampling levels leads to a faster decrease in training loss. Moreover, for shallow models ($\mathrm{lvl}\in\{1,2\}$), using two frames ($\mathtt{lh}=2$) yields a small reduction in loss compared with those using $\mathtt{lh}=3$, while for deeper models ($\mathrm{lvl}\in\{3,4\}$), three frames ($\mathtt{lh}=3$) perform slightly better than the $\mathtt{lh}=2$ counterparts. However, the difference at epoch 50 remains below $4\%$ and a longer history incurs extra computation and latency due to the need to wait for an additional future frame.
At 60 frames/s, each additional frame corresponds to approximately 16.7 ms of acquisition delay before inference can be performed. Since the performance improvement from lh=3 remained below 4\%, $\mathtt{lh}=2$ was retained to preserve lower latency under real-time constraints.
Based on this trade-off and the limitation given by the delay, we consider two input frames sufficient to distinguish moving foreground from static background, and accordingly fix $\mathtt{lh}=2$ in all subsequent experiments.

Finally, additional tests (not shown in the figure) indicate that the validation loss tends to stagnate around epoch 25, suggesting diminishing returns in model generalization despite the training loss continuing to decrease. This pattern is indicative of potential overfitting. Training was nevertheless performed for 50 epochs for consistency across experiments, while retaining the model corresponding to the best validation performance observed within the $50$-epoch training window.

\subsection{Trade-Off Analysis}\label{sec:tradeoff}
The second experiment is designed to identify the 2.5D U-Net model configuration that offers the best trade-off between segmentation performance and computational cost, under the practical constraint of real-time execution on a standard laptop.

\begin{figure}[!htbp]
  \centering
  \includegraphics[width=0.95\linewidth]{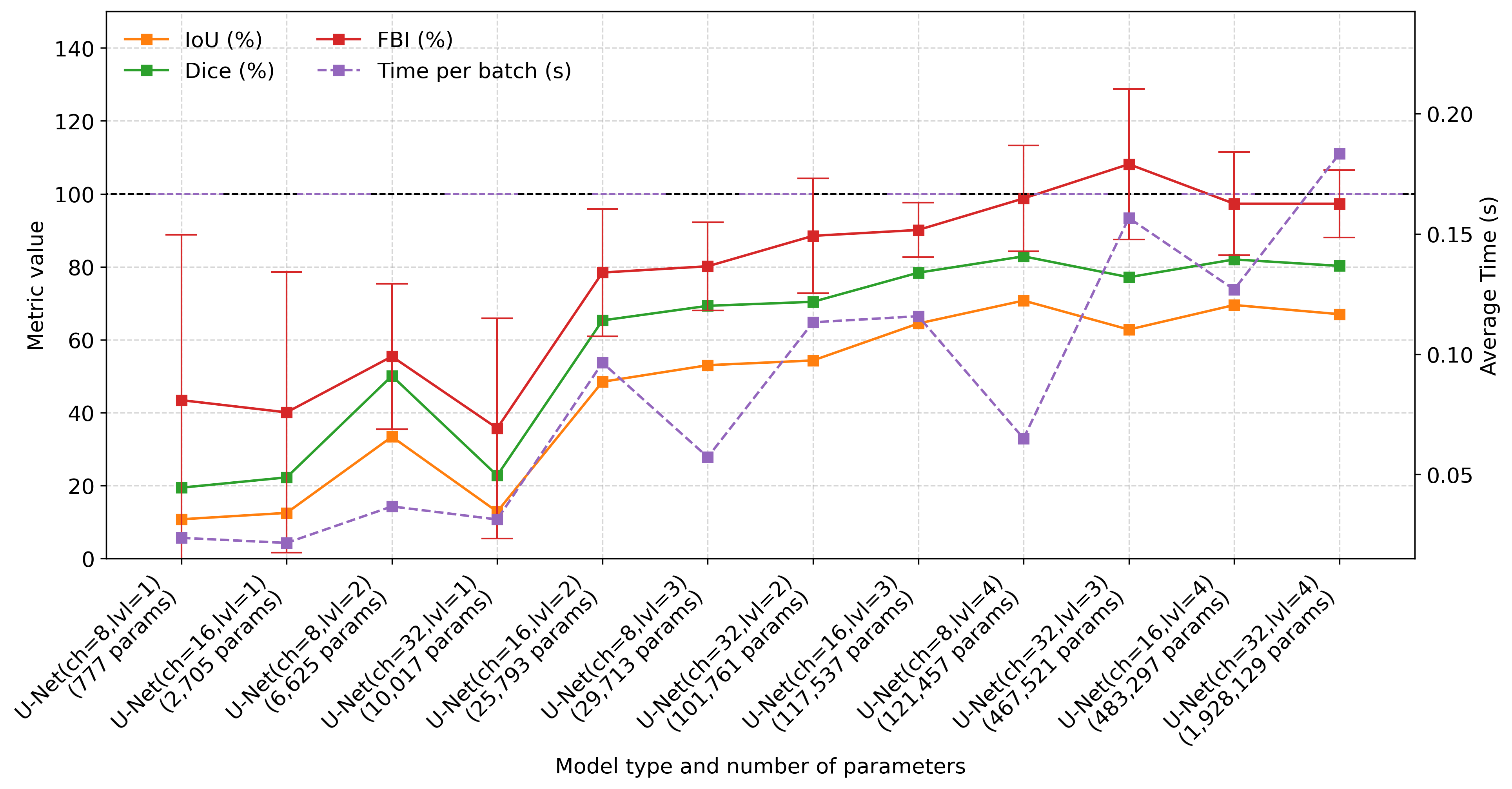}
  \caption{Segmentation performance (IoU, Dice, and FBI with standard deviation) and inference time per batch for different 2.5D U-Net architectures, evaluated with a fixed threshold of $t = 0.5$. Each model is identified by its configuration (channels, levels) and the corresponding number of parameters. FBI values are displayed with error bars to indicate variability across the test dataset. Inference time is plotted using a secondary y-axis.
  }
  \label{fig:acc_vs_params05}
\end{figure}

Figure~\ref{fig:acc_vs_params05} shows the performance of the tested 2.5D U-Nets with fixed threshold $t=0.5$ and $\mathtt{lh}=2$, while varying the number of base channels $\mathtt{ch} \in \{8, 16, 32\}$ and the number of downsampling levels $\mathtt{lvl} \in \{1, 2, 3, 4\}$. The performance is measured through the classical IoU, Dice, and FBI metrics (see Section~\ref{sec:metrics}), as well as average inference time per batch (right vertical axis), measured under real-time execution conditions. A single horizontal reference line guides interpretation by marking the target FBI value of 100$\%$ (indicating ideal coverage of the GME region) when read against the left axis, while simultaneously denoting the maximum acceptable inference time per batch when read against the right axis, beyond which real-time responsiveness may be compromised.

All metrics are computed over the entire test set.
Since accurate estimation of the overall GME burden is the primary objective of the proposed monitoring framework, FBI was considered the most clinically relevant calibration metric during model selection, while IoU and Dice were used to evaluate spatial segmentation consistency.
For this reason, we include error bars around the FBI values, computed as FBI$\pm\sigma_{10}$, where $\sigma_{10}$ is the standard deviation calculated by first grouping the test dataset into batches of size 10 images and then computing the FBI over each batch.

In Figure~\ref{fig:acc_vs_params05}, as expected, we observe that increasing the number of parameters generally leads to better performance, but higher inference times per batch. Thus, models with more than 100k parameters begin to achieve higher Dice and FBI values. However, beyond approximately 500k parameters, the performance gains tend to plateau. Meanwhile, the inference time per batch increases notably with model size, potentially introducing delays in real-time applications, especially on resource-constrained devices.\\
Since the real-time application aims to estimate the area of GME regions, we prioritize models with low FBI variability and a mean FBI possibly near 100$\%$. FBI values above 100$\%$ were preferred to FBI below 100$\%$ as discussed in Section~\ref{sec:metrics}, overestimating the GME area is preferable to underestimating it, as missing relevant regions may have more serious consequences. Based on this criterion, the most promising models are U-Net(ch=8, lvl=3,4), U-Net(ch=16, lvl=3), and U-Net(ch=32, lvl=4). While all four tend to slightly underestimate the GME area, probably due to the dominance of the background, the last configuration exceeds hardware constraints, making it less suitable for deployment.

A final operating parameter is the segmentation threshold $t$, which is applied to the predicted probability map after network inference. Lowering $t$ increases the number of pixels classified as GME, thereby generally increasing both TP and FP. Unlike the architectural parameters, $t$ is not learned during training and can be modified without retraining the network or changing its computational complexity. Based on the preliminary experiments, we fixed $t=0.4$ as a common operating point for the quantitative evaluation. The same threshold was subsequently applied to all compared architectures and kept fixed across all LOPO folds.

\begin{figure}[!htbp]
  \centering
  \includegraphics[width=0.95\linewidth]{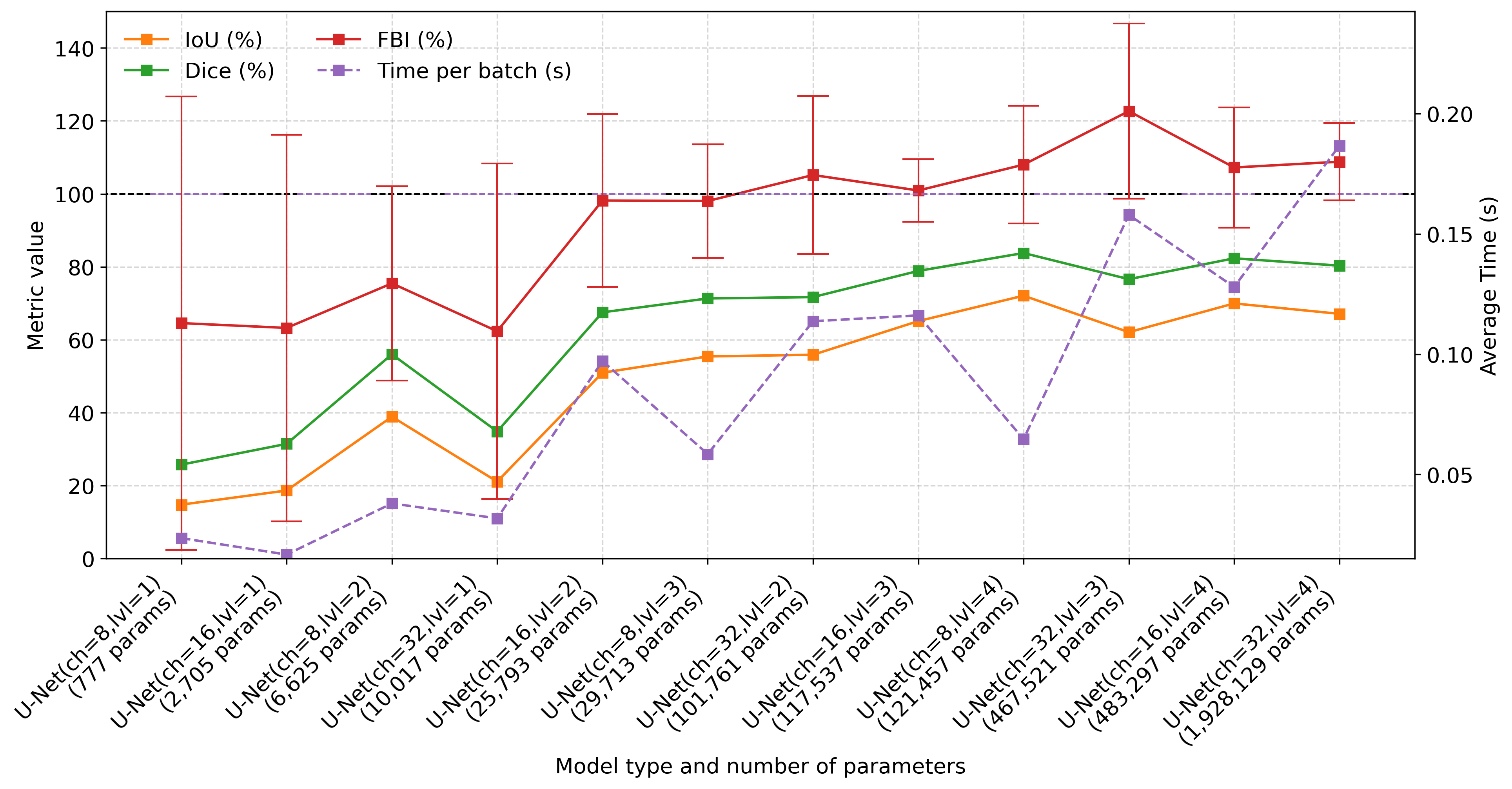}
  \caption{Segmentation performance (IoU, Dice, and FBI with standard deviation) and inference time per batch for different 2.5D U-Net architectures, evaluated with a threshold of $t = 0.4$. Each model is identified by its configuration and number of parameters.}\label{fig:acc_vs_params04}
\end{figure}

We note that the U-Net(ch=8, lvl=4) and U-Net(ch=16, lvl=3) models exhibit low FBI standard deviation (i.e., high reliability), moderate network size (i.e., faster inference), and strong overall performance. Both are therefore suitable for real-time applications. However, we focus on the latter, as its slightly lower FBI variance makes it the more reliable choice, while IoU and Dice (i.e., spatial agreement) remain comparable between the two models.

This adjustment confirms that the selected model offers the best trade-off between performance, reliability, parameter count, and real-time feasibility.

\subsection{Model Performance Evaluation}\label{sec:numres}
In this section, we present a comprehensive evaluation of the selected 2.5D U-Net model $\mathcal{U}$(ch=16, lvl=3,lh=2,t=0.4), demonstrating its effectiveness on real data.

After selecting the final architecture through the trade-off analysis described above, we further evaluated the selected model using a leave-one-patient-out protocol. The selected architecture and operating threshold were fixed before the LOPO evaluation and were kept identical across all folds. Each LOPO model was initialized and trained from scratch using only the corresponding training patients. This evaluation was designed to assess patient-level generalization within the available cohort and to compare the proposed lightweight 2.5D U-Net with additional state-of-the-art segmentation architectures, namely Attention U-Net (AttUNet)~\cite{Oktay2018AttentionUNet}, which introduces attention gates to focus the segmentation process on relevant image regions, and U-Net++~\cite{Zhou2018UNetPlusPlus}, which employs nested dense skip connections to improve feature fusion between encoder and decoder levels. The strict pixel-wise results of this comparison are reported in Table~\ref{tab:variant_metrics_0p40}. Because the clinical objective is bubble detection and embolic-load estimation rather than exact contour coincidence at uncertain boundaries, we also report in Table~\ref{tab:variant_metrics_0p40_radius} the radius-tolerant grace-zone evaluation defined in Section~\ref{sec:metrics}.

\begin{table}[!htbp]
\centering
\small
\setlength{\tabcolsep}{3.5pt}
\renewcommand{\arraystretch}{1.12}
\caption{Leave-one-patient-out segmentation performance (threshold $t=0.40$).}
\label{tab:variant_metrics_0p40}
\begin{adjustbox}{max width=\textwidth}
\begin{adjustbox}{max width=\textwidth}
\begin{tabular}{@{}lrrrrrrrrr@{}}
\toprule
Model & Params & IoU & Dice & AJI & Prec. & Rec. & FBI & ROC AUC \\
\midrule
AttUNet 2.5D (ch=16,lvl=3) & 494k & \textbf{\shortstack[r]{43.72\\{\scriptsize $\pm$ 10.21}}} & \textbf{\shortstack[r]{60.24\\{\scriptsize $\pm$ 9.69}}} & \textbf{\shortstack[r]{44.09\\{\scriptsize $\pm$ 11.23}}} & \shortstack[r]{61.10\\{\scriptsize $\pm$ 8.77}} & \textbf{\shortstack[r]{61.63\\{\scriptsize $\pm$ 15.65}}} & \textbf{\shortstack[r]{104.74\\{\scriptsize $\pm$ 27.44}}} & \shortstack[r]{99.47\\{\scriptsize $\pm$ 0.51}} \\
UNet++ 2.5D (ch=16,lvl=3) & 11.5M & \shortstack[r]{42.85\\{\scriptsize $\pm$ 12.24}} & \shortstack[r]{59.10\\{\scriptsize $\pm$ 11.95}} & \shortstack[r]{43.19\\{\scriptsize $\pm$ 13.28}} & \shortstack[r]{65.22\\{\scriptsize $\pm$ 12.28}} & \shortstack[r]{58.84\\{\scriptsize $\pm$ 18.81}} & \shortstack[r]{97.62\\{\scriptsize $\pm$ 37.29}} & \textbf{\shortstack[r]{99.63\\{\scriptsize $\pm$ 0.36}}} \\
UNet 2.5D (ch=16,lvl=3)$^\dagger$ & \textbf{118k} & \shortstack[r]{41.74\\{\scriptsize $\pm$ 12.11}} & \shortstack[r]{57.98\\{\scriptsize $\pm$ 12.28}} & \shortstack[r]{42.35\\{\scriptsize $\pm$ 13.75}} &\shortstack[r]{65.12\\{\scriptsize $\pm$ 10.77}} & \shortstack[r]{56.41\\{\scriptsize $\pm$ 18.50}} & \shortstack[r]{92.91\\{\scriptsize $\pm$ 35.55}} & \shortstack[r]{99.59\\{\scriptsize $\pm$ 0.42}} \\
UNet++ 2D (ch=32,lvl=4) & 13.9M & \shortstack[r]{40.42\\{\scriptsize $\pm$ 12.44}} & \shortstack[r]{56.62\\{\scriptsize $\pm$ 12.26}} & \shortstack[r]{40.85\\{\scriptsize $\pm$ 13.26}} &\textbf{\shortstack[r]{67.76\\{\scriptsize $\pm$ 12.12}}} & \shortstack[r]{50.19\\{\scriptsize $\pm$ 14.67}} & \shortstack[r]{78.84\\{\scriptsize $\pm$ 21.25}} & \shortstack[r]{99.38\\{\scriptsize $\pm$ 0.63}} \\
AttUNet 2D (ch=32,lvl=4) & 7.94M & \shortstack[r]{40.54\\{\scriptsize $\pm$ 13.95}} & \shortstack[r]{56.42\\{\scriptsize $\pm$ 14.65}} & \shortstack[r]{40.95\\{\scriptsize $\pm$ 13.81}} & \shortstack[r]{62.56\\{\scriptsize $\pm$ 7.65}} & \shortstack[r]{56.88\\{\scriptsize $\pm$ 22.92}} & \shortstack[r]{95.54\\{\scriptsize $\pm$ 39.86}} & \shortstack[r]{99.41\\{\scriptsize $\pm$ 0.47}} \\
AttUNet 2D (ch=32,lvl=1) & \textbf{103k} & \shortstack[r]{33.30\\{\scriptsize $\pm$ 14.79}} & \shortstack[r]{48.36\\{\scriptsize $\pm$ 16.61}} & \shortstack[r]{34.85\\{\scriptsize $\pm$ 13.99}} & \shortstack[r]{58.94\\{\scriptsize $\pm$ 9.22}} & \shortstack[r]{45.49\\{\scriptsize $\pm$ 23.47}} & \shortstack[r]{77.66\\{\scriptsize $\pm$ 40.10}} & \shortstack[r]{99.56\\{\scriptsize $\pm$ 0.40}} \\
\bottomrule
\end{tabular}
\end{adjustbox}
\end{adjustbox}
\end{table}

Table~\ref{tab:variant_metrics_0p40} shows that all evaluated architectures achieve comparable strict pixel-wise performance, but that generalization remains limited under the current low-patient regime. This suggests that the main limiting factor is the limited number of independent clinical acquisitions rather than the specific architectural choice. Within this setting, the proposed 2.5D U-Net remains close to the larger 2.5D variants of AttUNet and U-Net++, while still outperforming the 2D baselines despite using substantially fewer trainable parameters.

Table~\ref{tab:variant_metrics_0p40_radius} provides the complementary grace-zone evaluation, in which small localization offsets around the bubble boundary are tolerated. As expected, the overlap-based scores increase markedly for all models under this relaxed criterion, while the overall ranking remains essentially unchanged: the 2.5D models still perform best, Attention U-Net yields the highest tolerant overlap scores, and the proposed 2.5D U-Net remains close behind despite its much smaller size. This behavior indicates that a substantial part of the residual error in Table~\ref{tab:variant_metrics_0p40} is due to small boundary mismatches rather than complete failures to detect bubbles.

\begin{figure}[!htbp]
    \centering
    \includegraphics[width=0.650\linewidth]{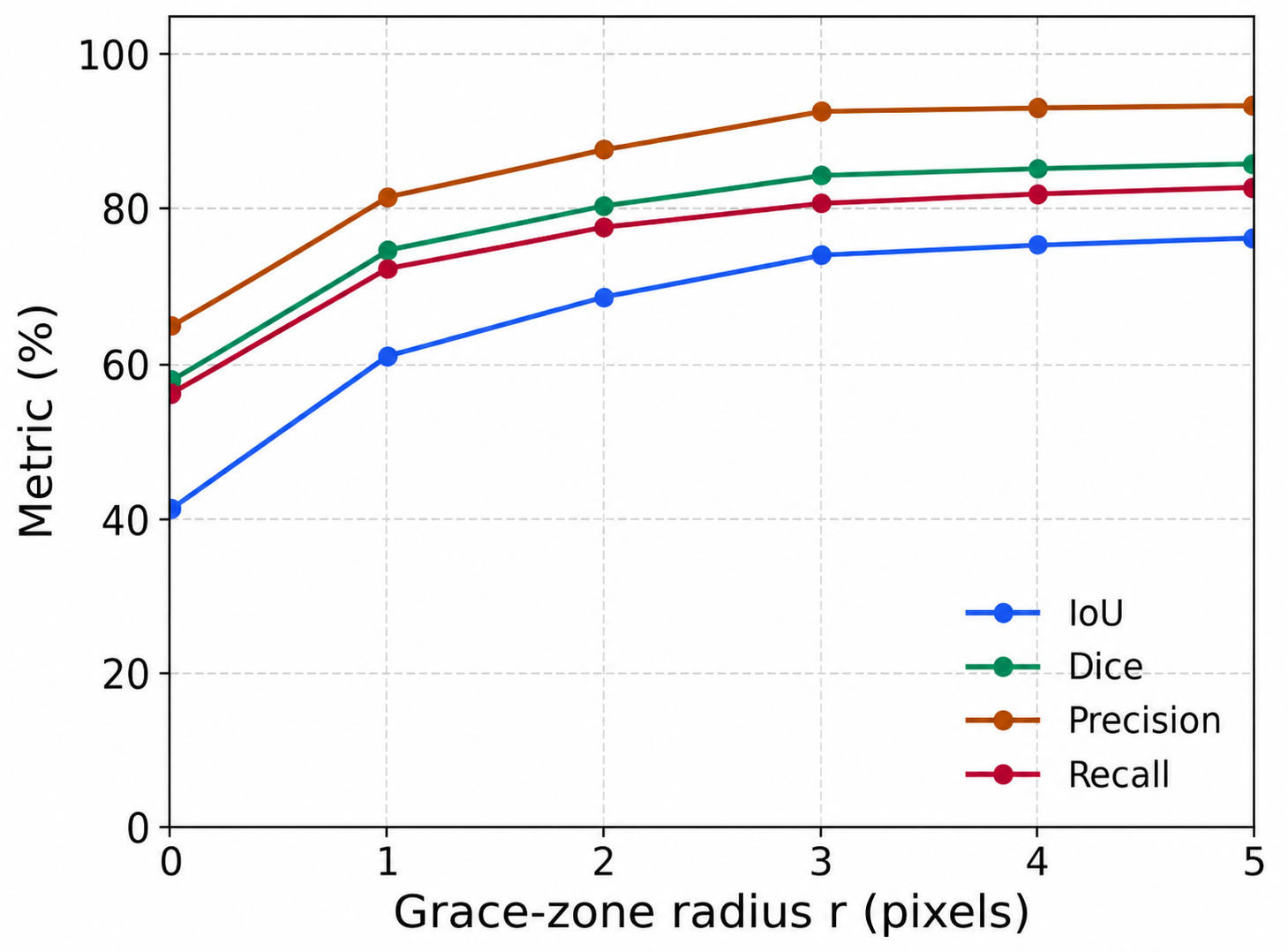}
    \caption{
    Sensitivity of the grace-zone radius $r$ for the selected 2.5D U-Net.
    }
    \label{fig:grace_zone_sensitivity}
\end{figure}

To assess the dependence of the radius-tolerant evaluation on the choice of $r$, we additionally performed a sensitivity analysis for the selected 2.5D U-Net using $r=0,\ldots,5$ pixels, where $r=0$ corresponds to the strict pixel-wise evaluation. The resulting mean IoU, Dice, precision, and recall values are shown in Fig.~\ref{fig:grace_zone_sensitivity}. The largest increase is observed when moving from the strict evaluation ($r=0$) to a tolerance of only one pixel ($r=1$). For larger radii, the metrics continue to improve more gradually and tend to stabilize from approximately $r=3$ onward.

The sensitivity analysis also provides an empirical context for the $r=3$ value used in the main radius-tolerant evaluation. Rather than representing a uniquely optimal or clinically defined threshold, $r=3$ lies in the region where the metric curves begin to plateau, providing a representative small image-space tolerance while avoiding substantially larger relaxation of the spatial criterion. Importantly, the same qualitative conclusion is already evident at smaller radii and therefore does not depend critically on the specific choice of $r=3$. Because the physical pixel size varies across patients according to acquisition settings and zoom level, a fixed radius in pixels does not correspond to the same physical distance in every examination. The three-pixel radius should therefore be interpreted only as an image-space tolerance for complementary offline evaluation and not as a universal physical distance or clinical safety margin.

The substantial difference between the strict and radius-tolerant results provides information about the spatial distribution of the segmentation errors. In particular, it suggests that the models frequently localize the correct GME regions, while a considerable proportion of the residual pixel-wise error is concentrated near their boundaries. The principal limitation therefore appears to concern precise contour delineation rather than complete failure to localize the embolic regions. Accordingly, the grace-zone results characterize the nature of the errors and complement the strict pixel-wise metrics.

We also note that additional tests, not reported in this paper, indicated that reducing the number of parameters in the 2D setting did not appear to improve the results. Importantly, while the difference in parameter count is substantial, the observed differences in both the strict and grace-zone evaluations remain relatively small. This finding is particularly relevant for the intended clinical use case, where real-time execution, low latency, and lightweight deployment are essential constraints. Therefore, although generalization remains limited with the present dataset, the proposed architecture represents a practical step toward real-time emboli detection and quantitative GME burden estimation. Larger multi-patient datasets will be required to fully assess generalization and to determine whether more complex architectures can provide consistent performance gains.

\begin{table}[!htbp]
\centering
\small
\setlength{\tabcolsep}{3.5pt}
\renewcommand{\arraystretch}{1.12}
\caption{Leave-one-patient-out segmentation performance (threshold $t=0.40$) under the radius-tolerant grace-zone criterion ($r=3$ pixels).}
\label{tab:variant_metrics_0p40_radius}
\begin{tabular}{@{}lrrrrrr@{}}
\toprule
Model & Params & IoU & Dice  & Prec. & Rec. \\
\midrule
AttUNet 2.5D (ch=16,lvl=3) & 494k & \textbf{\shortstack[r]{77.58\\{\scriptsize $\pm$ 14.09}}} & \textbf{\shortstack[r]{86.73\\{\scriptsize $\pm$ 9.31}}}  & \shortstack[r]{92.44\\{\scriptsize $\pm$ 6.85}} & \textbf{\shortstack[r]{82.66\\{\scriptsize $\pm$ 11.18}}} \\
UNet++ 2.5D (ch=16,lvl=3) & 11.50M & \shortstack[r]{76.92\\{\scriptsize $\pm$ 17.63}} & \shortstack[r]{85.90\\{\scriptsize $\pm$ 12.12}} & \textbf{\shortstack[r]{93.77\\{\scriptsize $\pm$ 6.14}}} & \shortstack[r]{82.47\\{\scriptsize $\pm$ 15.86}} \\
UNet 2.5D (ch=16,lvl=3)$^\dagger$ & \textbf{118k} & \shortstack[r]{73.95\\{\scriptsize $\pm$ 15.69}} & \shortstack[r]{84.13\\{\scriptsize $\pm$ 11.34}} &  \shortstack[r]{92.55\\{\scriptsize $\pm$ 4.55}} & \shortstack[r]{80.54\\{\scriptsize $\pm$ 15.72}} \\
UNet++ 2D (ch=32,lvl=4) & 13.89M & \shortstack[r]{70.24\\{\scriptsize $\pm$ 17.93}} & \shortstack[r]{81.33\\{\scriptsize $\pm$ 12.98}} & \shortstack[r]{92.19\\{\scriptsize $\pm$ 12.91}} & \shortstack[r]{75.24\\{\scriptsize $\pm$ 13.83}} \\
AttUNet 2D (ch=32,lvl=4) & 7.94M & \shortstack[r]{71.06\\{\scriptsize $\pm$ 21.56}} & \shortstack[r]{81.28\\{\scriptsize $\pm$ 16.44}} & \shortstack[r]{93.32\\{\scriptsize $\pm$ 4.60}} & \shortstack[r]{75.70\\{\scriptsize $\pm$ 19.41}} \\
AttUNet 2D (ch=32,lvl=1)& \textbf{103k} & \shortstack[r]{64.88\\{\scriptsize $\pm$ 24.30}} & \shortstack[r]{78.60\\{\scriptsize $\pm$ 18.17}} & \shortstack[r]{90.85\\{\scriptsize $\pm$ 9.72}} & \shortstack[r]{74.13\\{\scriptsize $\pm$ 20.59}} \\
\bottomrule
\end{tabular}
\end{table}

\begin{figure}[!htbp]
    \centering
    \subfloat[Real-time processing workflow.]{%
        \includegraphics[width=\linewidth]{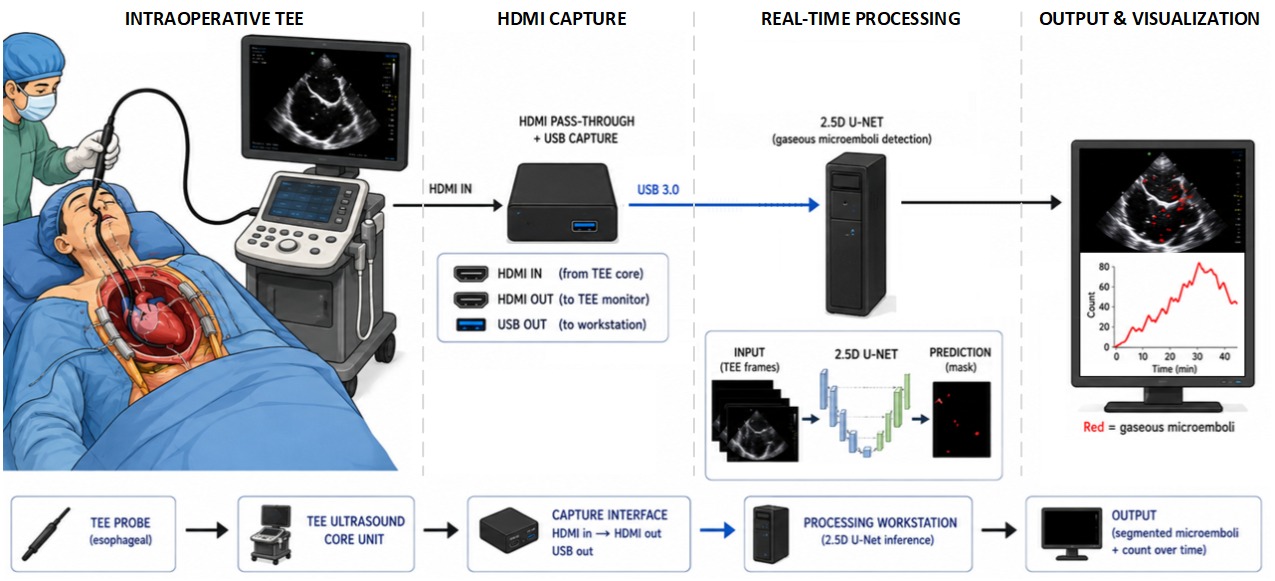}%
    }

    \medskip

    \subfloat[Example segmentation output.]{%
        \includegraphics[width=0.7\linewidth]{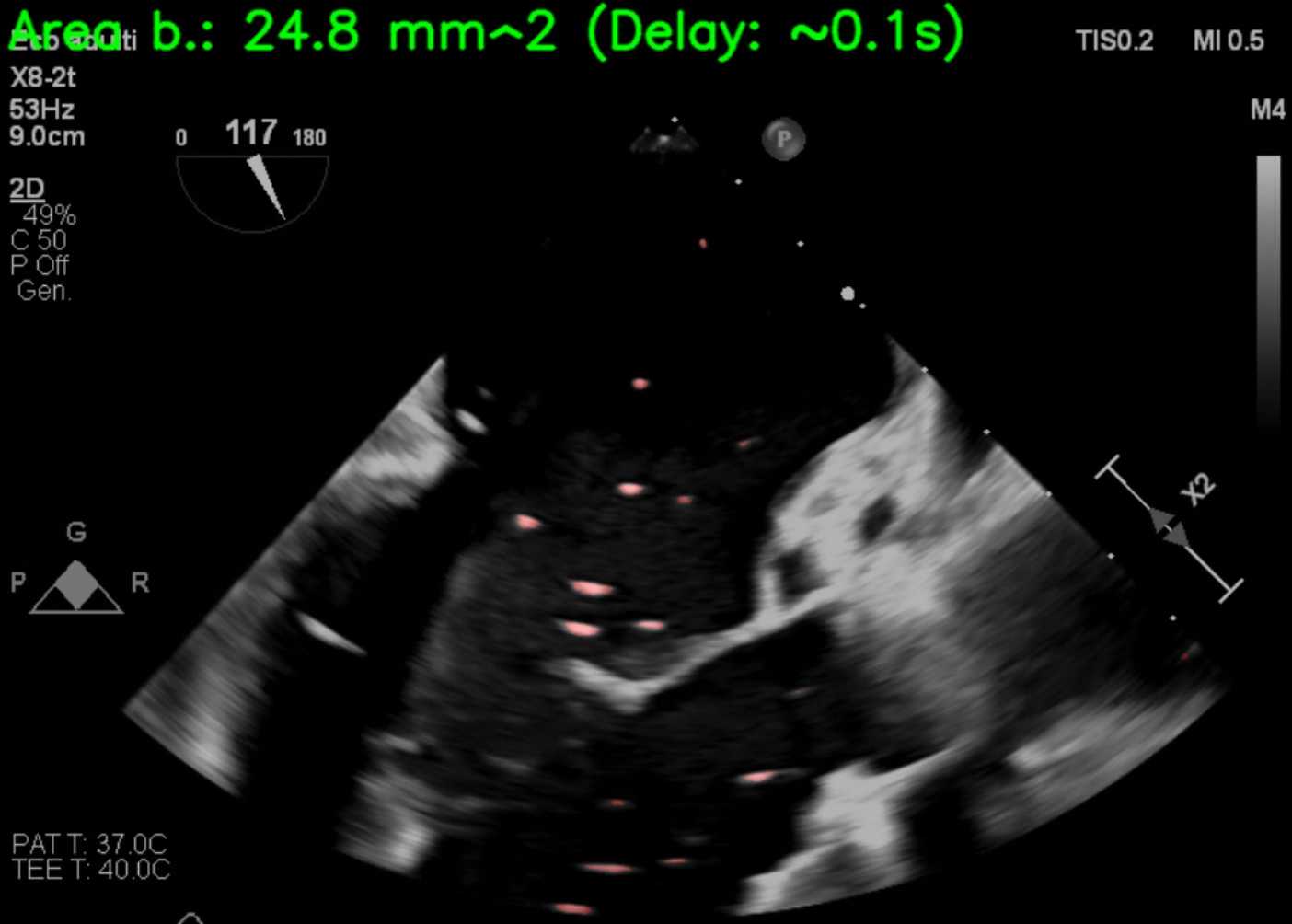}%
    }
    \caption{Real-time intraoperative GME monitoring framework. (a) Schematic representation of the complete processing pipeline: the intraoperative TEE video stream is mirrored through the capture interface, transferred via USB 3.0 to a local processing workstation, analyzed by the 2.5D U-Net, and returned as a real-time visualization of the detected GME regions and their quantitative evolution. (b) Representative frame acquired during real-time operation. Red overlays indicate GMEs detected by the selected 2.5D U-Net model $\mathcal{U}(ch=16,lvl=3,lh=2,t=0.4)$, while the estimated GME area is displayed at the top of the image.}
    \label{fig:realtime_segmentation}
\end{figure}

\subsection{Evaluation on GME-negative dataset}\label{S5:negative_dataset}

To validate the proposed system on an external dataset and complement the leave-one-patient-out evaluation, we analyzed the publicly available SR-TEE dataset, which contains transesophageal echocardiographic sequences without visible GMEs~\cite{Xu2023SRTEE}. The dataset comprises 4394 frames from real TEE cases and 15,825 synthetically generated TEE frames, for a total of 20,219 frames. These data were completely unseen during model development and include cardiac views acquired or generated at different probe positions and zoom levels from those represented in our training dataset.

This experiment was designed as a negative-control test to determine whether the model produces spurious detections when applied to ordinary cardiac structures and echocardiographic background patterns from an independent source. Since no visible GMEs are present in these sequences, we assumed a completely black ground-truth mask for every frame. As a consequence, every pixel classified as GME was considered a false-positive pixel.

\begin{figure}[!htbp]
\centering
\includegraphics[width=0.9\linewidth]{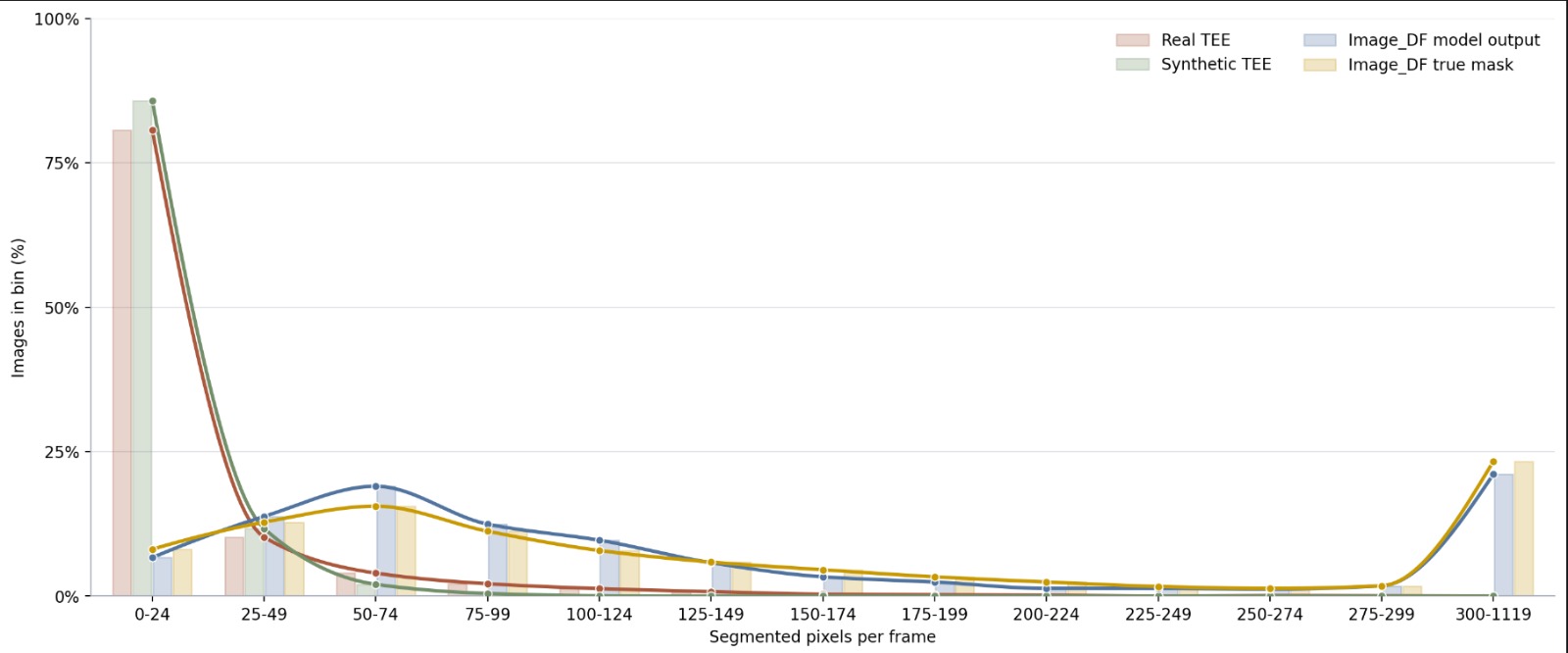}
\caption{Distributions of the number of pixels classified as GME on the GME-positive dataset (blue), its corresponding ground-truth masks (yellow), and the predictions on the real (red) and synthetic (green) subsets of the GME-negative SR-TEE dataset.}
\label{fig:distribution}
\end{figure}

To contextualize the negative-control results, we compared these outputs with those obtained on a GME-positive reference subset of the original dataset summarized in Table~\ref{tab:dataset_summary}. This reference subset includes only frames whose ground-truth masks contain at least one GME-positive pixel. Figure~\ref{fig:distribution} therefore compares four frame-wise distributions: the model predictions and corresponding ground-truth masks on this GME-positive subset, and the model predictions on the real and synthetic subsets of the GME-negative SR-TEE dataset.
The distributions obtained on both GME-negative subsets are strongly concentrated near zero and are clearly separated from those observed on the GME-positive recordings.  In particular, more than 90\% of the predictions on the GME-negative data contain fewer than 50 false-positive pixels, an area approximately comparable to that of two average-sized manually segmented GMEs. Moreover, the model produced completely empty masks, containing no false-positive pixels, for 50\% of the real TEE frames and 29\% of the synthetic TEE frames. The median false-positive area was also only 5 pixels on the synthetic TEE subset, indicating that at least half of these frames contained no more than five false-positive pixels. This value is markedly lower than the medians observed on the GME-positive dataset, which were 95 pixels for the model outputs and 105 pixels for the ground-truth masks, i.e., approximately 100 pixels in both cases. More generally, both the predicted and ground-truth distributions on our GME-positive dataset are shifted toward substantially larger values, showing that the network produces a markedly stronger response when GMEs are present. This separation is also reflected in the maximum pixel counts: spurious predictions reached at most 315 pixels on the real TEE subset and 125 pixels on the synthetic subset. In contrast, the GME-positive distributions exhibit substantially heavier upper tails, with pixel counts approaching or exceeding 1000 pixels.  These results indicate that the model does not systematically misclassify normal cardiac anatomy or echocardiographic background patterns as GME. Nevertheless, this negative-control experiment should not be interpreted as a substitute for validation on an independent GME-positive cohort, but as a complementary assessment of model specificity in the absence of visible emboli.

\subsection{Real-Time Segmentation}\label{S5.4:realtime}
Beyond the offline quantitative evaluation, the selected 2.5D U-Net was integrated into a real-time inference pipeline to assess its suitability for intraoperative use. This setup was designed not only to provide immediate visual feedback on the presence and spatial distribution of GMEs, but also to estimate their visible area during the intervention.
The prototype was physically implemented and tested as an external real-time processing system connected to the Philips Epiq echocardiography unit through an HDTV USB 3.0 video-capture device.

Figure~\ref{fig:realtime_segmentation}(a) summarizes the complete intraoperative processing workflow. The TEE video signal is mirrored through the capture interface so that the conventional echocardiographic visualization remains available to the clinical team, while the same stream is transferred to a local workstation for 2.5D U-Net inference. The resulting segmentation is then displayed as an additional real-time output without modifying the standard TEE acquisition workflow. Figure~\ref{fig:realtime_segmentation}(b) shows a representative output generated in real time by the implemented system on a single frame captured during surgery.

At the top of the output screen, the GME area (Area b.) value is displayed, representing the estimated overall GME area. This estimate is computed by counting the number of pixels classified as positive (ones) in the network’s output mask, corresponding to the numerator of the FBI quantity.
The pixel count is then rescaled to a physical measurement in \(\textrm{mm}^2\), taking into account the zoom level selected by the doctor operating the echocardiography system. The segmented area is computed as the product of the number of segmented pixels and the known physical area represented by a single pixel under the current acquisition settings. This provides a real-time quantitative assessment of the area visible during the intervention.

For visualization purposes, given that the video runs at 60 frames per second, we apply an exponential moving average (EMA), with $\alpha=\frac{1}{2}$, and update the GME area estimate every ten frames to ensure a readable, accurate, and interpretable value. In addition, Figure~\ref{fig:realtime_segmentation} reports an estimate of the processing delay, which becomes relevant when the available hardware does not fully support real-time performance. Specifically, this value is computed from the number of frames currently waiting in the processing queue and their corresponding acquisition time. Therefore, values below $0.2 \mathrm{s}$ indicate that the system is operating, without accumulating additional delay due to inference or frame acquisition. Displaying this delay helps verify correct pipeline operation and allows the surgical team to assess whether latency remains within acceptable bounds.
This measurement can be used during the procedure as a real-time indicator of GME presence and extent, and its reliability can be assessed directly by comparing the predicted GMEs with the live visual feedback.
The quantitative characterization of GME may also be relevant beyond their total area. Experimental and numerical studies have shown that both the size and number of GMEs can influence infarction burden, with smaller gaseous microemboli being associated with a greater total volume of cerebral infarctions for a given air volume \cite{Schaefer1036,Stehouwer2020}. These findings highlight that individual GME size may provide complementary information to the total embolic-area estimate produced by the proposed framework.
Moreover, in the postoperative phase, the network can be applied to provide the same quantitative information almost instantaneously, transforming the original data into a more interpretable format that may also be accessible to non-clinical personnel.

\section{Conclusions}
This work identified a lightweight 2.5D U-Net architecture as a practical and computationally efficient approach for real-time detection and quantification of gaseous microemboli in intraoperative cardiac ultrasound. By leveraging short-term temporal context, the proposed model improves the discrimination between moving emboli and static cardiac background structures while preserving low-latency execution compatible with intraoperative deployment.

A leave-one-patient-out comparison against Attention U-Net and U-Net++ further showed that all evaluated architectures experience limited generalization under the current low-data regime.
This suggests that the primary limitation is the reduced number of independent clinical acquisitions rather than the specific architecture itself. Nevertheless, the proposed 2.5D U-Net achieved segmentation performance comparable to substantially larger models while requiring significantly fewer trainable parameters, which is particularly relevant for lightweight real-time clinical applications.
U-Net-based architectures have shown strong adaptability across medical imaging tasks~\cite{NEHA2025100416}.
Although our study was conducted on a limited number of subjects using a single imaging device, the negligible predicted GME burden on the independent GME-negative dataset provides preliminary evidence that the proposed model does not systematically generate spurious detections in the absence of visible emboli. Whether the proposed protocol generalizes to other centers with similar acquisition settings remains to be evaluated on independent GME-positive cohorts.

Overall, the framework represents a step toward real-time AI-assisted monitoring of embolic load, with potential benefits for intraoperative decision-making and patient safety.

\paragraph*{Acknowledgments} The authors acknowledge the support of Fondazione Aldo e Cele Daccò through the 'Fondo EOC-USI'. The funding was essential for carrying out the present research.

\bibliographystyle{unsrtnat}
\bibliography{references}

@article{ChungEM,
    doi = {10.1371/journal.pone.0122166},
    author = {Chung, Emma M. L. AND Banahan, Caroline AND Patel, Nikil AND Janus, Justyna AND Marshall, David AND Horsfield, Mark A. AND Rousseau, Clément AND Keelan, Jonathan AND Evans, David H. AND Hague, James P.},
    journal = {PLOS ONE},
    publisher = {Public Library of Science},
    title = {Size Distribution of Air Bubbles Entering the Brain during Cardiac Surgery},
    year = {2015},
    month = {04},
    volume = {10},
    url = {https://doi.org/10.1371/journal.pone.0122166},
    pages = {1-11},
    number = {4},
}

@article{Kihara2021,
  author       = {Kihara, Kazuki and Orihashi, Kazumasa},
  title        = {Investigation of air bubble properties: Relevance to prevention of coronary air embolism during cardiac surgery},
  journal      = {Artificial Organs},
  volume       = {45},
  number       = {9},
  pages        = {E349-E358},
  year         = {2021},
  month        = {September},
  doi          = {10.1111/aor.13975},
  url          = {https://doi.org/10.1111/aor.13975},
  publisher    = {International Center for Artificial Organs and Transplantation and Wiley Periodicals, Inc.},
  issn         = {1525-1594},
  pmid         = {33908061},
  language     = {English},
  note         = {© 2021 International Center for Artificial Organs and Transplantation and Wiley Periodicals, Inc.}
}

@article{Orihashi2024,
  author       = {Kazumasa Orihashi and Tsuyoshi Miyata},
  title        = {Retained intracardiac air in cardiovascular surgery: a re-visited problem},
  journal      = {General Thoracic and Cardiovascular Surgery},
  volume       = {72},
  number       = {7},
  pages        = {429--438},
  year         = {2024},
  month        = {July},
  doi          = {10.1007/s11748-024-02041-x},
  url          = {https://doi.org/10.1007/s11748-024-02041-x},
  issn         = {1863-6713},
  note         = {Published online 2024/07/01}
}

@article{Kurz2022,
  author = {Kurz, A. and Hauser, K. and Mehrtens, H. A. and Krieghoff-Henning, E. and Hekler, A. and Kather, J. N. and Fröhling, S. and von Kalle, C. and Brinker, T. J.},
  title = {Uncertainty estimation in medical image classification: Systematic review},
  journal = {JMIR Medical Informatics},
  volume = {10},
  number = {8},
  pages = {e36427},
  year = {2022},
  doi = {10.2196/36427}
}

@inproceedings{Long_2015_CVPR,
  author    = {Long, Jonathan and Shelhamer, Evan and Darrell, Trevor},
  title     = {Fully Convolutional Networks for Semantic Segmentation},
  booktitle = {Proceedings of the IEEE Conference on Computer Vision and Pattern Recognition (CVPR)},
  year      = {2015}
}

@InProceedings{He_2017_ICCV,
author = {He, Kaiming and Gkioxari, Georgia and Dollar, Piotr and Girshick, Ross},
title = {Mask R-CNN},
booktitle = {Proceedings of the IEEE International Conference on Computer Vision (ICCV)},
month = {October},
year = {2017}
}

@InProceedings{ronneberger2015unet,
author="Ronneberger, Olaf and Fischer, Philipp and Brox, Thomas",
editor="Navab, Nassir and Hornegger, Joachim and Wells, William M. and Frangi, Alejandro F.",
title="U-Net: Convolutional Networks for Biomedical Image Segmentation",
booktitle="Medical Image Computing and Computer-Assisted Intervention -- MICCAI 2015",
year="2015",
publisher="Springer International Publishing",
address="Cham",
pages="234--241",
isbn="978-3-319-24574-4"
}

@article{Antonello2023,
author = {Antonello, Paola and Morone, Diego and Pirani, Edisa and Uguccioni, Mariagrazia and Thelen, Marcus and Krause, Rolf and Pizzagalli, Diego},
year = {2023},
month = {01},
pages = {},
title = {Tracking unlabeled cancer cells imaged with low resolution in wide migration chambers via U-NET class-1 probability (pseudofluorescence)},
volume = {17},
journal = {Journal of Biological Engineering},
doi = {10.1186/s13036-022-00321-9}
}

@misc{Aircatch,
  author = {Angino, A. and Trotti, K.},
  title = {{Aircatch}},
  howpublished = {\url{https://github.com/AnginoA/Aircatch}},
  year = {2025},
  note = {GitHub repository}
}

@article{Ershov2022,
  author       = {Dmitry Ershov and Minh-Son Phan and Joanna W. Pylvänäinen and Stéphane U. Rigaud and Laure Le Blanc and Arthur Charles-Orszag and James R. W. Conway and Romain F. Laine and Nathan H. Roy and Daria Bonazzi and Guillaume Duménil and Guillaume Jacquemet and Jean-Yves Tinevez},
  title        = {TrackMate 7: integrating state-of-the-art segmentation algorithms into tracking pipelines},
  journal      = {Nature Methods},
  year         = {2022},
  volume       = {19},
  number       = {7},
  pages        = {829--832},
  doi          = {10.1038/s41592-022-01507-1},
  url          = {https://doi.org/10.1038/s41592-022-01507-1}
}

@article{Unet25D,
  author       = {Christoph Angermann and
                  Markus Haltmeier and
                  Ruth Steiger and
                  Sergiy Pereverzyev Jr. and
                  Elke Ruth Gizewski},
  title        = {Projection-Based 2.5D U-net Architecture for Fast Volumetric Segmentation},
  journal      = {CoRR},
  volume       = {abs/1902.00347},
  year         = {2019},
  url          = {http://arxiv.org/abs/1902.00347},
  eprinttype    = {arXiv},
  eprint       = {1902.00347},
  bibsource    = {dblp computer science bibliography, https://dblp.org}
}

@ARTICLE{Unet2,
  author={Siddique, Nahian and Paheding, Sidike and Elkin, Colin P. and Devabhaktuni, Vijay},
  journal={IEEE Access}, 
  title={U-Net and Its Variants for Medical Image Segmentation: A Review of Theory and Applications}, 
  year={2021},
  volume={9},
  number={},
  pages={82031-82057},
  doi={10.1109/ACCESS.2021.3086020}}

@article{arbelaez2011contour,
  title   = {Contour Detection and Hierarchical Image Segmentation},
  author  = {Arbel{\'a}ez, Pablo and Maire, Michael and Fowlkes, Charless and Malik, Jitendra},
  journal = {IEEE Transactions on Pattern Analysis and Machine Intelligence},
  volume  = {33},
  number  = {5},
  pages   = {898--916},
  year    = {2011},
  doi     = {10.1109/TPAMI.2010.161}
}

@inproceedings{cheng2021boundary,
  title     = {Boundary IoU: Improving Object-Centric Image Segmentation Evaluation},
  author    = {Cheng, Bowen and Girshick, Ross and Doll{\'a}r, Piotr and Berg, Alexander C. and Kirillov, Alexander},
  booktitle = {Proceedings of the IEEE/CVF Conference on Computer Vision and Pattern Recognition},
  pages     = {15334--15342},
  year      = {2021},
  doi       = {10.1109/CVPR46437.2021.01508}
}

@article{nikolov2021clinically,
  title   = {Clinically Applicable Segmentation of Head and Neck Anatomy for Radiotherapy: Deep Learning Algorithm Development and Validation Study},
  author  = {Nikolov, Stanislav and Blackwell, Sam and Zverovitch, Alexei and Mendes, Ruheena and Livne, Michelle and De Fauw, Jeffrey and Patel, Yojan and Meyer, Clemens and Askham, Harry and Romera-Paredes, Bernadino and Kelly, Christopher and Karthikesalingam, Alan and Chu, Carlton and Carnell, Dawn and Boon, Cheng and D'Souza, Derek and Moinuddin, Syed Ali and Garie, Bethany and McQuinlan, Yasmin and Ireland, Sarah and Hampton, Kiarna and Fuller, Krystle and Montgomery, Hugh and Rees, Geraint and Suleyman, Mustafa and Back, Trevor and Hughes, C{\'i}an Owen and Ledsam, Joseph R. and Ronneberger, Olaf},
  journal = {Journal of Medical Internet Research},
  volume  = {23},
  number  = {7},
  pages   = {e26151},
  year    = {2021},
  doi     = {10.2196/26151}
}

@misc{Xu2023SRTEE,
  author = {Xu, Jialang and Jin, Yueming and Martin, Bruce and Smith, Andrew and Wright, Susan and Stoyanov, Danail and Mazomenos, Evangelos B.},
  title = {{SR-TEE Dataset for Regressing Simulation to Real: Unsupervised Domain Adaptation for Automated Quality Assessment in Transoesophageal Echocardiography}},
  howpublished = {\url{https://doi.org/10.5522/04/23699736.v2}},
  year = {2023},
  publisher = {University College London},
  doi = {10.5522/04/23699736.v2},
  note = {Dataset}
}

@article{Oktay2018AttentionUNet,
  title={Attention U-Net: Learning Where to Look for the Pancreas},
  author={Oktay, Ozan and Schlemper, Jo and Folgoc, Loic Le and Lee, Matthew and Heinrich, Mattias and Misawa, Kazunari and Mori, Kensaku and McDonagh, Steven and Hammerla, Nils Y. and Kainz, Bernhard and Glocker, Ben and Rueckert, Daniel},
  journal={arXiv preprint arXiv:1804.03999},
  year={2018}
}

@inproceedings{Zhou2018UNetPlusPlus,
  title={{UNet++}: A Nested {U-Net} Architecture for Medical Image Segmentation},
  author={Zhou, Zongwei and Siddiquee, Md Mahfuzur Rahman and Tajbakhsh, Nima and Liang, Jianming},
  booktitle={Deep Learning in Medical Image Analysis and Multimodal Learning for Clinical Decision Support},
  pages={3--11},
  year={2018},
  publisher={Springer}
}

@article {Schaefer1036,
	author = {Schaefer, Tabea C and Greive, Svenja and Bierwisch, Claas and Mohseni-Mofidi, Shoya and Heiland, Sabine and Kramer, Martin and M{\"o}hlenbruch, Markus A and Bendszus, Martin and Vollherbst, Dominik F},
	title = {Iatrogenic air embolism: influence of air bubble size on cerebral infarctions in an experimental in vivo and numerical simulation model},
	volume = {16},
	number = {10},
	pages = {1036--1041},
	year = {2024},
	doi = {10.1136/jnis-2023-020739},
	publisher = {British Medical Journal Publishing Group},
	issn = {1759-8478},
	URL = {https://jnis.bmj.com/content/16/10/1036},
	eprint = {https://jnis.bmj.com/content/16/10/1036.full.pdf},
	journal = {Journal of NeuroInterventional Surgery}
}

@article{Stehouwer2020,
  author  = {Stehouwer, Marco C. and de Vroege, Roel and Bruggemans, Eline F. and Hofman, Frederik N. and Molenaar, Meyke A. and van Oeveren, Wim and de Mol, Bastian A. and Bruins, Peter},
  title   = {The influence of gaseous microemboli on various biomarkers after minimized cardiopulmonary bypass},
  journal = {Perfusion},
  year    = {2020},
  volume  = {35},
  number  = {3},
  pages   = {202--208},
  doi     = {10.1177/0267659119867572},
  pmid    = {31402782}
}

@article{NEHA2025100416,
title = {An analytics-driven review of U-Net for medical image segmentation},
journal = {Healthcare Analytics},
volume = {8},
pages = {100416},
year = {2025},
issn = {2772-4425},
doi = {https://doi.org/10.1016/j.health.2025.100416},
url = {https://www.sciencedirect.com/science/article/pii/S2772442525000358},
author = {Fnu Neha and Deepshikha Bhati and Deepak Kumar Shukla and Sonavi Makarand Dalvi and Nikolaos Mantzou and Safa Shubbar},
}

\end{document}